\documentclass[journal]{IEEEtran}
\ifCLASSINFOpdf
   \usepackage[pdftex]{graphicx}
\else
\fi

\usepackage{amssymb}
\usepackage{caption}
\usepackage{comment}
\usepackage{multirow}
\usepackage[dvipsnames]{xcolor}

\usepackage{array}
\newcolumntype{P}[1]{>{\centering\arraybackslash}p{#1}}
\begin{document}
%
\title{Oriented Bounding Boxes for Small and Freely Rotated Objects}
%
%
%

\author{Mohsen~Zand,~\IEEEmembership{Member,~IEEE,}
        Ali~Etemad,~\IEEEmembership{Senior Member,~IEEE,}
        and~Michael~Greenspan,~\IEEEmembership{Member,~IEEE}
\thanks{M. Zand is with the Department of Electrical and Computer Engineering, Queen's University, Kingston, ON, Canada, and Ingenuity Labs Research Institute, Queen's University, Kingston, ON, Canada e-mail: (m.zand@queensu.ca).}
\thanks{A. Etemad and M. Greenspan are with the Department of Electrical and Computer Engineering, Queen's University, Kingston, ON, Canada, and Ingenuity Labs Research Institute, Queen's University, Kingston, ON, Canada.}}

%
%

\markboth{IEEE Transactions on Geoscience and Remote Sensing, VOL. XX, NO. XX, XXXX, 2021}
{Zand \MakeLowercase{\textit{et al.}}: Object Detection as Multi-Scale Semantic Segmentation for Remote Sensing Imagery}
%



\maketitle

\begin{abstract}
A novel object detection method is presented that handles freely rotated objects of arbitrary sizes, including tiny
objects as small as $2 \times 2$ pixels. Such tiny objects appear frequently in remotely sensed images, and present a challenge to recent object detection algorithms. 
More importantly, current object detection methods have been designed
originally to accommodate axis-aligned bounding box detection, and therefore fail to accurately localize oriented boxes
that best describe
freely rotated objects. In contrast, the proposed CNN-based approach uses potential pixel information at multiple scale levels without the need for any external resources, such as anchor boxes. The method encodes the precise location and orientation of features of the target objects at grid cell locations. Unlike existing methods which regress the bounding box location and dimension, the proposed method learns all the required information by classification, which has the added benefit of enabling oriented bounding box detection without any extra computation. It thus infers the bounding boxes only at inference time by finding the minimum surrounding box for every set of the same predicted class labels.
Moreover, a rotation-invariant feature representation is applied to each scale, which imposes a regularization constraint to enforce covering the 360 degree range of in-plane rotation of the training samples to share similar features. Evaluations on the xView and DOTA datasets show that the proposed method uniformly improves performance over existing state-of-the-art methods. 
\end{abstract}


%
\IEEEpeerreviewmaketitle

\section{Introduction}
%
%
%
%
\IEEEPARstart{O}{bject} detection in remotely sensed imagery, as acquired traditionally by satellites and more recently by drones, has become increasingly important with the rapid expansion and increased  diversity of practical real-world applications~\cite{liu2016ship,kalantar2017multiple,li2020object}.  Satellite images in particular contain high visual diversity and artifacts due to  natural occurrences such as varied lighting and weather conditions, seasonal variations of foliage, clouds and cloud shadows over landscapes, 
as well as human-induced variations such as the existence of vehicles and built environments~\cite{usman2019weakly,khan2017forest}. Given these challenges, the analysis of scenes from remotely sensed images for different applications is a problem of significance and a challenging and current research direction. 

Despite the significant advances over the past few years in research related to object detection~\cite{ren2015faster,redmon2018yolov3} and semantic segmentation~\cite{girshick2014rich,kemker2018algorithms} using deep learning methods, a number of challenges remain for remotely sensing imagery.
For example, object localization and classification are particularly difficult due to the abundance and highly dense scenes with tiny and densely clustered objects~\cite{li2020object}. The high resolution imagery that has recently become available in satellite images enables the detection of smaller and smaller objects. Nonetheless, localizing and detecting objects that are sometimes only \textit{a few pixels in size} is still non-trivial. 

Another challenge is that compared to many common datasets, which naturally constrain object orientations, the objects in remotely sensed images can be oriented at any arbitrary in-plane rotation~\cite{van2018you}. For example, trees are usually oriented vertically in dash-cam images, whereas ships can be pointed in any direction when they are imaged from overhead satellites. Since learned features in object detection algorithms are based on training examples, object instances at different rotation angles must be present in the training dataset. Otherwise, the learned features may not be sufficiently discriminating, which can result in lower accuracy in subsequent object detection and recognition tasks. 
    
    
    A further challenge is that it is often desirable that freely rotated objects be detected in their \textit{oriented bounding boxes}. While in standard object detection approaches ground truths are labeled by axis-aligned (e.g., horizontal or vertical) bounding boxes, focusing on rotated bounding boxes (e.g., \cite{xia2018dota,azimi2018towards,yang2019scrdet}) are recently becoming popular in remote sensing applications due to their ability to more precisely localize the objects and provide object orientation within the image plane.

To summarize, the proposed method is specifically motivated to address the localization and detection of objects under the following two challenging conditions, which occur routinely in remotely sensed images:
\begin{itemize}
    \item Objects are of various sizes, from very large (e.g. buildings) to very small (e.g. automobiles); 
    \item  Objects are oriented at any arbitrary in-plane rotation. 
\end{itemize}
A further motivation is to detect a distinct oriented bounding box around each detected object since these multi-scale, small and arbitrary oriented objects cannot be detected without misalignment between bounding boxes and objects and axis-aligned boxes usually contain the other objects.

\begin{figure*}[!t]
\begin{center}
   \includegraphics[width=0.8\linewidth]{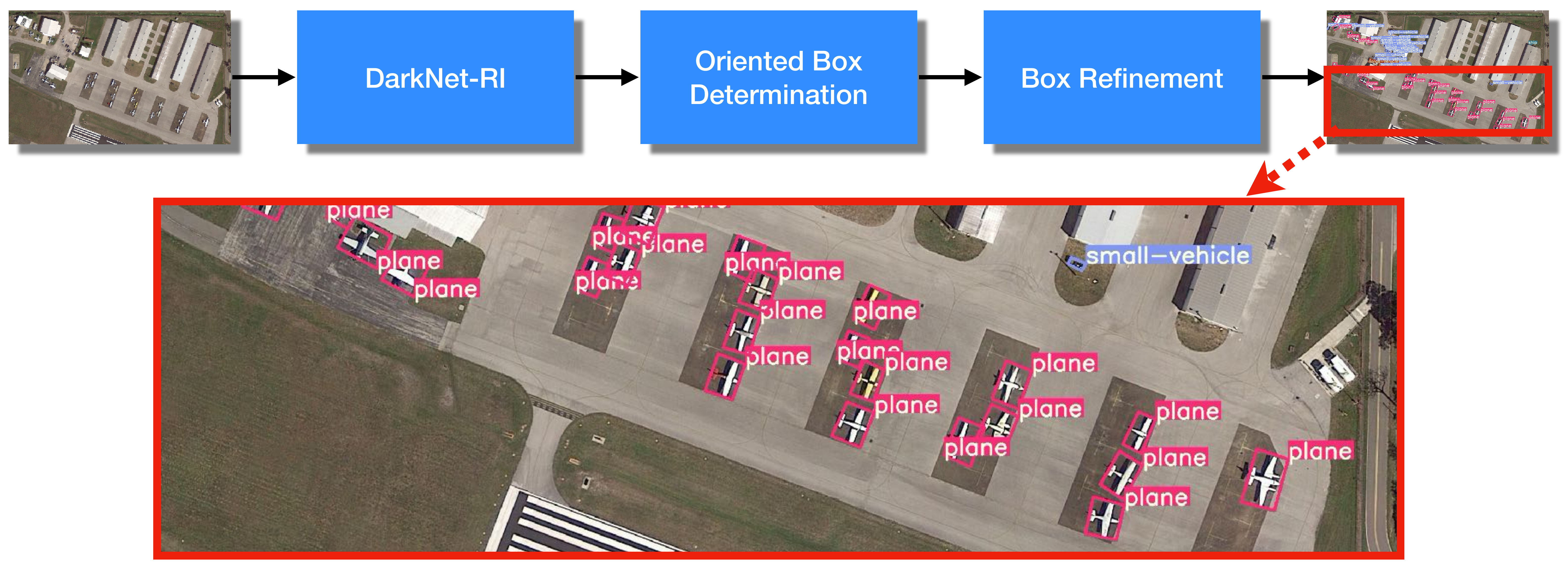}
\end{center}
   \caption{An overview of the proposed method.}
\label{fig:model_overview}
\end{figure*}

In this paper, we address these challenges and present a novel CNN-based object detection method tailored for \textit{small and freely in-plane rotated objects} with applications to remotely sensed imagery. Our goal is to create a pipeline capable of detecting both axis-aligned and oriented bounding boxes (OBB).  Our approach uses a custom CNN based on the DarkNet-53 backbone architecture~\cite{redmon2018yolov3}, to extract multi-scale, rotation-invariant representations from input images. 
We call this network \emph{DarkNet-RI} (where RI stands for Rotation-Invariant),
the main processing elements of which are presented in Fig.~\ref{fig:model_overview}.

Following representation extraction, an \textit{oriented box determination module} uses morphological operations to remove noise and artifacts, and more importantly  determines the oriented bounding boxes. Next, a \textit{box refinement module} is used to further perform non-maximum suppression (NMS)~\cite{neubeck2006efficient} to remove redundant and overlapping bounding boxes in order to generate the final output.
As illustrated in Fig.~\ref{fig:model}, our DarkNet-RI uses a 5-level pyramidal structure for dense and multi-scale feature extraction, and includes a rotation-invariant layer that enforces a regularization constraint during the representation learning stage.

The three main contributions of this work are as follows:
\begin{enumerate}
    \item 
We propose a novel object detection approach for remote sensing imagery. The method is capable of detecting boxes that are oriented with their respective enclosed objects as opposed to most other localization approaches~\cite{zhang2019scale,zhuang2019single,qiu2019a2rmnet} which detect horizontally or vertically aligned boxes only. As compared to existing oriented bounding box detection~\cite{azimi2018towards,li2018multiscale, yang2019scrdet}, our method is a pure classification approach without extra computation for oriented bounding boxes.
\item
We show that our network, DarkNet-RI, is a novel solution for multi-scale and rotation-invariant representation learning. 
\item
Our extensive experiments illustrate that our method outperforms other general object detection networks namely YOLOv3~\cite{redmon2018yolov3} and SSD~\cite{liu2016ssd}, as well as networks specifically designed for remote sensing, namely YOLT~\cite{van2018you} and Reduced focal loss (RFL)~\cite{sergievskiy2019reduced} in the axis-aligned bounding box detection task. It also obtains outstanding performance in the oriented bounding box detection task in comparison with state-of-the-art methods, namely R-DFPN~\cite{yang2018automatic}, RRPN~\cite{ma2018arbitrary}, ICN~\cite{azimi2018towards}, SCRDet~\cite{yang2019scrdet}, and ROI Transformer~\cite{ding2019learning}.
\end{enumerate}

The remainder of this paper is organized as follows: Section 2 reviews the related work of both general and specifically designed object detection methods used for remote sensing applications. Our approach is detailed in Section 3, followed by extensive experimental evaluations in Section 4. Finally, Section 5 summarizes the work and outlines a number of future research directions. 

 

\section{Related Work}
The automatic detection and classification of objects is a fundamental task in computer vision, and while well-explored in the literature, it remains challenging both in general and for certain specific problem scenarios. In the following, we first give an overview of common object detection methods, and then review  object detection approaches specifically designed for remote sensing imagery.

\begin{table*}[h]
\caption{A summary of object detection methods in remote sensing imaging.}
\label{table:related_work} 
\centering
\scriptsize
    \begin{tabular}{|P{0.75cm}|P{2.5cm}|P{3cm}|P{1.3cm}|P{2cm}|P{5cm}|}
    \hline
        \textbf{Method} & \textbf{Architecture} & \textbf{Dataset} & \textbf{Aligned vs. Oriented} & \textbf{OBB Strategy} & \textbf{Contributions} \\
    \hline\hline
        \cite{liu2016ship} & A two-cascaded model followed by binary linear programming & HRSC2016 collected from Google Earth~\cite{liu2017high} & Oriented & Additional regression for rotation angle & Potential candidates are selected from a ship rotated bounding box space \\
    \hline
        \cite{li2017rotation} & RPN and Local-contextual feature fusion network & NWPU VHR-10~\cite{cheng2016learning} & Aligned & - & Additional multiangle anchors are added into RPN in the Faster R-CNN \\
    \hline
        \cite{guo2018geospatial} & Multi-scale object proposal and detection networks & NWPU VHR-10 & Aligned & - & Multi-scale anchor boxes are added to multi-scale feature maps\\
    \hline
        \cite{azimi2018towards} & Image cascade network (ICN) using ResNet~\cite{he2016deep} & DOTA~\cite{xia2018dota}, NWPU VHR-10, and UCAS-AOD~\cite{zhu2015orientation} & Oriented & Predicting four corner points of the OBB & An image cascade network aggregated with FPN to extract features for regressing the offsets of OBBs relative to axis-aligned boxes.\\
    \hline
        \cite{van2018you} & DarkNet & COWC~\cite{mundhenk2016large} & Aligned & - & A finer-grained network and data augmentation \\
    \hline
        \cite{li2018multiscale} & RPN & 640 images collected from Google Earth & Oriented & Additional regression for rotation angle & An angle parameter is added to the original four parameters for OBB \\
    \hline
        \cite{yang2018automatic} & Rotation Dense Feature Pyramid Networks & 1000 images collected from Google Earth & Oriented & Additional regression for rotation angle & A multi-scale rotation region detection using Rotation Region Proposal Networks (RRPNs) and rotation anchors to detect ships. \\
    \hline
        \cite{zhuang2019single} & Multi-scale feature fusion detector & NWPU VHR-10, and RSD-GOD & Aligned & - & A multi-scale feature fusion \\
    \hline
        \cite{zhang2019cad} & Faster R-CNN with FPNs & DOTA, and NWPU VHR10 & Oriented & Additional OBB regression branch & Using contextual regions by multi-scale co-occurrence features and/or co-occurrence objects surrounding the objects of interest. \\
    \hline
        \cite{ding2019learning} & ResNet101 & DOTA, and HRSC2016 & Oriented & Additional regression for rotation angle & Spatial transformations are applied on RoIs and parameters are learned under the supervision of OBB annotations. \\
    \hline
        \cite{wang2019sard} & ResNet101 & DOTA, and HRSC2016 & Oriented & Additional regression for rotation angle & A 
        scale-aware rotated object detector (SARD) 
        using a RPN and feature fusion module added to the R-CNN. \\
    \hline
        \cite{sergievskiy2019reduced} & RPN and feature pyramid network & xView~\cite{lam2018xview} & Aligned & - & A reduced focal loss to address imbalanced nature of datasets  \\
    \hline
        \cite{wei2019oriented} & 104-Hourgalss~\cite{law2018cornernet} followed by CornerNet~\cite{law2018cornernet} & DOTA & Oriented & OBB middle lines & An anchor-free method to predict a pair of middle lines to detect each object. \\
    \hline
        \cite{qiu2019a2rmnet} & Adaptively aspect ratio multi-scale network (A2RMNet) & DOTA, NWPU VHR-10, RSOD~\cite{long2017accurate}, and UCAS-AOD & Aligned & - & Multi-scale feature gate fusion and aspect ratio attention networks based on RoIs \\
    \hline
        \cite{zhang2019scale} & Scale adaptive proposal network (SAPNet) & DOTA, and NWPU VHR-10 & Aligned & - & A multiobject detection CNN based on R-CNN and a scale adaptive proposal network \\
    \hline
        \cite{yang2019scrdet} & ResNet101 & DOTA, and NWPU VHR-10 & Oriented & Additional regression for rotation angle & Multi-layer features are fused with anchor sampling, and a supervised pixel attention network is used. \\
    \hline
        \cite{zhou2020objects} & ResNet101 & DOTA, NWPU VHR-10, and UCAS-AOD & Oriented & Predicting four corner points of the OBB in the polar coordinate system &
        Representing the bounding boxes in the polar coordinate system, and predicting the center point and regressing one polar radius and two polar angles. \\
    \hline
    \end{tabular}
\end{table*}

\subsection{General Object Detection Methods}
The objective is that each detected object be assigned a semantic class label, such as \emph{building} or \emph{road}, along with a bounding box which delineates its location. Both the label and the location of the image objects must therefore be predicted simultaneously. Instead of investigating every image pixel for the presence of each object, region-based CNN (R-CNN) algorithms~\cite{girshick2014rich,girshick2015fast,ren2015faster} propose investigating a limited set of boxes (i.e. regions) within the input image for general-purpose object detection. In these methods, a set of region candidates are first generated, and then object class labels and their bounding boxes are obtained using classification and regression tasks. The most popular region-based models include R-CNN~\cite{girshick2014rich}, Fast R-CNN~\cite{girshick2015fast}, and Faster R-CNN~\cite{ren2015faster},
and they have produced state-of-the-art results on many standard datasets. For instance, Faster R-CNN achieved a mean average precision (\emph{mAP}) of $70.4\%$ on the PASCAL VOC 2012 dataset.

The introduction of R-CNN by Girshick~\textit{et al.}~\cite{girshick2014rich} led to rich and high-level convolutional feature extraction using CNNs. It used about 2000 region proposals for each image, from which deep features of these proposals were extracted and fed into a set of SVMs. Image objects were finally labeled as either background or other object classes, and a linear regressor was utilized to localize each detected object in a bounding box. 

The region proposal computations were inefficiently repeated in R-CNN, which motivated the proposal of Fast R-CNN~\cite{girshick2015fast} to share the computations of CNN features for all region proposals. Fast R-CNN employed a region of interest (RoI) layer over the entire image to run the CNN-model only one time, which resulted in a faster, more efficient execution than R-CNN. To explore more subtle region proposal generation, Ren~\textit{et al.}~\cite{ren2015faster} proposed  leveraging a region proposal network (RPN) into the Fast R-CNN framework to obtain the end-to-end Faster R-CNN architecture. It utilized an RPN to generate several anchor boxes in different aspect ratios and sizes. The RPN takes convolutional feature maps as its input and applies a sliding window over these feature maps to predict anchor boxes at each window as region proposals. The convolutional features are then extracted from these proposals. For each anchor box, an object label and a bounding box regressor are predicted.

Another successful CNN-based object detection system is YOLO~\cite{redmon2016you} and  its two variants YOLO9000~\cite{redmon2017yolo9000} and YOLOv3~\cite{redmon2018yolov3}, which predict bounding boxes and associated classes directly using a unified CNN framework. In YOLO, the coordinates of the bounding boxes are predicted using fully connected layers on top of the convolutional feature extractors. In YOLOv3, however, the fully connected layer is removed and anchor boxes are used instead for predicting bounding boxes. It utilizes the RPN for anchor box prediction at each grid cell. 
In the default YOLOv3 method, images are downsampled by a factor of 32 and used in a 13$\times$13 prediction grid. This means that even with a carefully-designed anchor box set, two objects with their centroids separated by less than a 32 pixel distance cannot be differentiated or thus detected. These methods however impose constraints on the number of predictable bounding boxes in any grid cell. In particular, they rely on external region proposal systems to localize the objects in the image. These strict spatial constraints on the bounding boxes can therefore result in unreliable features for detecting objects, specifically small objects in dense areas. In spite of their success in standard datasets, for this reason these methods struggle with  overhead satellite images that may contain a variety of tiny and densely clustered objects. 

\begin{figure*}[t]
\begin{center}
   \includegraphics[width=0.95\linewidth]{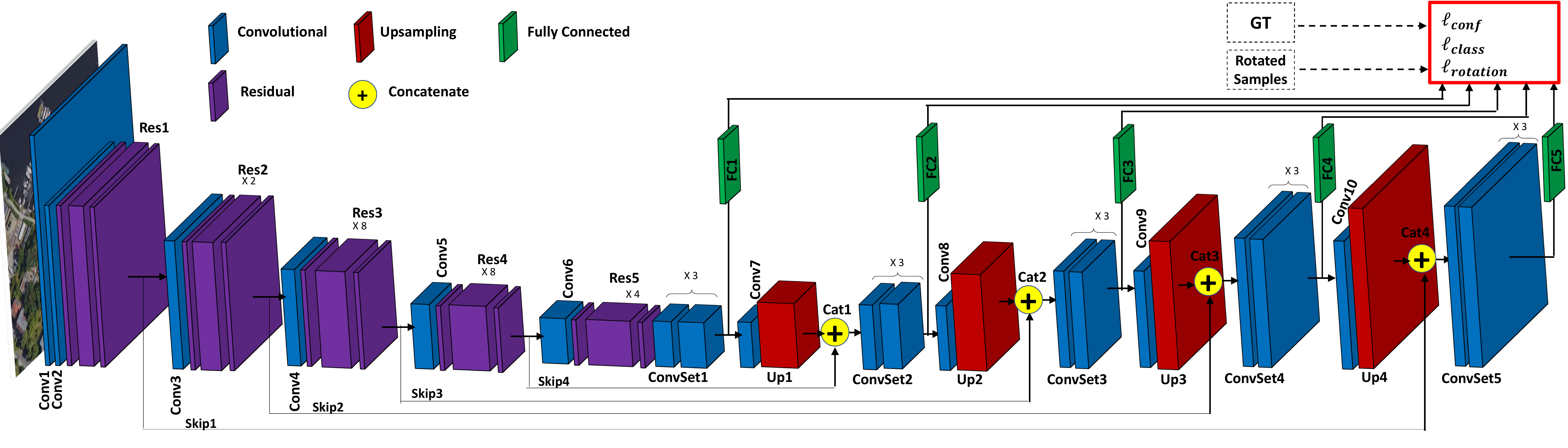}
\end{center}
   \caption{The main architecture of the DarkNet-RI.}
\label{fig:model}
\end{figure*}

\subsection{Remote Sensing Object Detection}
Object detection in remotely sensed images has been extensively studied since the 1980s~\cite{cheng2016survey}. Since common object detection methods fail to overcome the challenges of the remote sensing images, specialized CNN-based object detection methods have been recently developed to specifically tackle their issues~\cite{li2017rotation,zhuang2019single,li2020object}. An overview of these methods is given in Table~\ref{table:related_work}. 

When applied to remotely sensed images, CNN-based methods have demonstrated success at detecting objects such as ships, airplanes, and buildings, although they still lack reliability and robustness at detecting smaller objects, especially those that are densely clustered. These CNN-based methods, nevertheless, are still based on the existing general-purpose object detectors. Li~\textit{et al.}~\cite{li2017rotation}, for example, proposed leveraging an RPN with additional multi-angle anchor boxes. Guo~\textit{et al.}~\cite{guo2018geospatial} proposed to use a multi-scale object proposal network and trained an object detector on the obtained proposals using another multi-scale object detection network. Similarly, Zhang~\textit{et al.}~\cite{zhang2019scale} adopted a multilayer region proposal network to generate multiscale object proposals. They showed the improvement on the remote sensing images by feature fusion from two different layers. Zhuang~\textit{et al.}~\cite{zhuang2019single} designed a single shot detection model with multi-scale feature fusion. They extracted low-level and high-level features using up-sampling and concatenation blocks in different feature maps and aggregated for the final prediction results. Cheng~\textit{et al.}~\cite{cheng2020cross} proposed a cross-scale feature fusion (CSFF) framework based on the Faster R-CNN~\cite{ren2015faster} and Feature Pyramid Network (FPN)~\cite{lin2017feature}.

Although detecting oriented bounding boxes has often been overlooked in the literature, there are few methods which addressed this issue in the remote sensing literature. Liu~\textit{et al.}~\cite{liu2016ship} proposed to use nearly closed-form ship rotated bounding box space and selected highly potential candidates from this space by binary linear programming. They scored latent candidates by a two-cascaded linear model. Li~\textit{et al.}~\cite{li2018multiscale} used five parameters to regress location-angle offset based on the multiscale rotated prior boxes for ship detection in remote sensing images. To learn the ship features, they searched the best-matched prior boxes with the ground-truth boxes to obtain the positive samples. Yang~\textit{et al.}~\cite{yang2018automatic} proposed a multi-scale rotation region detection method which adopted Rotation Region Proposal Networks (RRPNs) and used rotation anchors to detect ships in their respective orientation. The RRPN was initially introduced by Ma~\textit{et al.}~\cite{ma2018arbitrary} 
for arbitrary-oriented text detection. It generated proposals with the object orientation information and regressed the offsets of OBBs relative to oriented proposals. 
Azimi~\textit{et al.}~\cite{azimi2018towards} proposed an image cascade network (ICN) aggregated with feature pyramid network to extract features for regressing the offsets of OBBs relative to axis-aligned boxes. The features were fed into rotation-based region proposal and region of interest networks to generate predictions. Yang~\textit{et al.}~\cite{yang2019scrdet} proposed an OBB detector for small, cluttered and rotated objects, namely SCRDet. They used a sampling fusion network
which combined multi-layer feature with anchor sampling for small object detection. RoI Transformer was presented by Ding~\textit{et al.}~\cite{ding2019learning} which applied spatial transformations on RoIs and learn the transformation parameters under the supervision of OBB annotations. Zhang~\textit{et al.}~\cite{zhang2019cad} utilized Faster R-CNN with FPNs and learned scene global context and object local context using a global context network (GCN) and a pyramid local context network (PLCNet), respectively. In particular, they proposed to use contextual regions using multi-scale co-occurrence features and/or co-occurrence objects surrounding the objects of interest. Moreover, they detected oriented boxes using an additional OBB regression branch to their network. Wei~\textit{et al.}~\cite{wei2019oriented} proposed an anchor-free method similar to FCOS~\cite{tian2019fcos}, which predicted a pair of middle lines corresponding to the interior object lines to locate each object. Wang~\textit{et al.}~\cite{wang2019sard} proposed a 
scale-aware rotated object detector (SARD) which leveraged a feature fusion module in the R-CNN. It used a normalization strategy for OBB representation. In~\cite{zhou2020objects}, arbitrary-oriented objects were detected by predicting the center point and regressing one polar radius and two polar angles. It represented the four corner points of the OBB in the polar coordinate system. 

These methods mostly rely on regression-based approaches which often suffer from ambiguity in the definition of learning targets~\cite{yang2018multi,liu2020rgb}.
Our method however is a novel framework for remote sensing object detection which significantly offers improvements on the existing methods.

\begin{table}[h]
\caption{DarkNet-RI Structure}
\label{table:conv_detail} 
\centering
\small
\begin{tabular}{cccccc}
  \hline
  Label in & \# of Output & Filter & \multirow{2}{*}{Stride} & Output & \# of \\
  Fig.~\ref{fig:model} & Channels & Size & & Size & Layers\\
  \hline\hline
  Conv1  & 32 & 3$\times$3	& 1	& 512$\times$512 & 1 \\
 \hline
Conv2   & 64	& 3$\times$3	& 2	& 256$\times$256 &  1 \\
\hline
\multirow{2}{*}{Res1}  & 32	& 1$\times$1	& 1	& 	\multirow{2}{*}{256$\times$256}& \multirow{2}{*}{1}
\\
  & 64	& 3$\times$3	& 1	&  & 
\\
\hline
Conv3   & 128	& 3$\times$3	& 2	& 128$\times$128	& 1
\\
\hline
\multirow{2}{*}{Res2}  & 64 & 1$\times$1	& 1	& \multirow{2}{*}{128$\times$128}	& \multirow{2}{*}{2}
\\
  & 128 & 3$\times$3	& 1	& 	&	\\
\hline
Conv4   & 256	& 3$\times$3	& 2	& 64$\times$64 &  1 \\
\hline
\multirow{2}{*}{Res3}  & 128 & 1$\times$1	& 1	& \multirow{2}{*}{64$\times$64}	& \multirow{2}{*}{8}
\\
& 256 & 3$\times$3	& 1	& 	&	\\
\hline
Conv5   & 512	& 3$\times$3	& 2	& 32$\times$32 &  1 \\
\hline
\multirow{2}{*}{Res4}  & 256 & 1$\times$1	& 1	& \multirow{2}{*}{32$\times$32}	& \multirow{2}{*}{8}
\\
  & 512 & 3$\times$3	& 1	& 	&	\\
\hline
Conv6   & 1024	& 3$\times$3	& 2	& 16$\times$16 &  1 \\
\hline
\multirow{2}{*}{Res5}  & 512 & 1$\times$1	& 1	& \multirow{2}{*}{16$\times$16}	& \multirow{2}{*}{4}
\\
  & 1024 & 3$\times$3	& 1	& 	&	\\
\hline
\multirow{2}{*}{ConvSet1}  & 512 & 1$\times$1	& 1	& \multirow{2}{*}{16$\times$16}	& \multirow{2}{*}{3}
\\
  & 1024 & 3$\times$3	& 1	& 	& 	\\
\hline
FC1 & ${C\!+\!1}$ & -	& -	& $1\!\!\times\!\!(C\!+\!1)$	& 1
\\
\hline
Conv7  & 256	& 1$\times$1	& 1	& 16$\times$16 &  1 \\
\hline
Up1  & 256	& -	& 2	& 32$\times$32 &  1 \\
\hline
Cat1  & 768	& -	& -	& 32$\times$32 &  1 \\
\hline
\multirow{2}{*}{ConvSet2} & 256 & 1$\times$1	& 1	& \multirow{2}{*}{32$\times$32}	& \multirow{2}{*}{3}
\\
 & 512 & 3$\times$3	& 1	& 	& \\
\hline
FC2 & ${C\!+\!1}$ & -	& -	& $1\!\!\times\!\!(C\!+\!1)$	& 1
\\
\hline
Conv8  & 128	& 1$\times$1	& 1	& 32$\times$32 &  1 \\
\hline
Up2  & 128	& -	& 2	& 64$\times$64 &  1 \\
\hline
Cat2  & 384	& -	& -	& 64$\times$64 &  1 \\
\hline
\multirow{2}{*}{ConvSet3} & 128 & 1$\times$1	& 1	& \multirow{2}{*}{64$\times$64}	& \multirow{2}{*}{3} 
\\
 & 256 & 3$\times$3	& 1	& 	&	\\
\hline
FC3 & ${C\!+\!1}$ & -	& -	& $1\!\!\times\!\!(C\!+\!1)$	& 1
\\
\hline
Conv9  & 128	& 1$\times$1	& 1	& 64$\times$64 &  1 \\
\hline
Up3  & 128	& -	& 2	& 128$\times$128 &  1 \\
\hline
Cat3  & 256	& -	& -	& 128$\times$128 &  1 \\
\hline
\multirow{2}{*}{ConvSet4} & 128 & 1$\times$1	& 1	& \multirow{2}{*}{128$\times$128}	&\multirow{2}{*}{3} 
\\
 & 256 & 3$\times$3	& 1	& 	& 	\\
\hline
FC4 & ${C\!+\!1}$ & -	& -	& $1\!\!\times\!\!(C\!+\!1)$	& 1
\\
\hline
Conv10 & 128	& 1$\times$1	& 1	& 128$\times$128 &  1 \\
\hline
Up4 & 128	& -	& 2	& 256$\times$256 &  1 \\
\hline
Cat4 & 192	& -	& -	& 256$\times$256 &  1 \\
\hline
\multirow{2}{*}{ConvSet5} & 128 & 1$\times$1	& 1	& \multirow{2}{*}{256$\times$256}	&\multirow{2}{*}{3} 
\\
 & 256 & 3$\times$3	& 1	& 	&	\\
\hline
FC5 & ${C\!+\!1}$ & -	& -	& $1\!\!\times\!\!(C\!+\!1)$	& 1
\\
\hline
\end{tabular}
\caption*{Conv$X$ = A convolutional operation,
Res$X$ = Residual Block,
ConvSet$X$ = Convolution Block (including two convolutional operations),
FC$X$ = Fully Connected Layer,
Up$X$ = Upsampling Layer,
Cat$X$ = Contactenation,
$C$ = \# of Classes.}
\end{table}

\section{Proposed Method}
\subsection{Problem Setup}
Let input image $I \in \mathbb{N}^{W \times H}$ of width and height  $W$ and $H$ respectively contain a set of objects $O=\{o_1, o_2, ... , o_n\}$. For each object $o_i$, an in-plane rotation $R_i$ describes the orientation of its corresponding bounding box $b_i = (x_1,y_1,x_2,y_2,x_3,y_3,x_4,y_4,c_i)$, where $c_i$ is the $o_i$ class label, $o_i$ is inscribed within $b_i$, and the major axes of $b_i$ and $o_i$ are parallel. Our goal in this paper is to detect $b_i$ and $c_i$ for each corresponding $o_i$, with $R_i$ implicitly resolved by the coordinates of $b_i$. 

Our model consists of three major components, multi-scale semantic segmentation, oriented box determination, and box refinement, 
which are described in detail in the following sections.


\subsection{Multi-Scale Semantic Segmentation}
We use DarkNet-53~\cite{redmon2018yolov3} as the backbone of our proposed network,
which we call \emph{DarkNet-RI},
given its success in object detection and widespread use in solutions such as YOLT~\cite{van2018you} and multi-scale feature fusion detector~\cite{zhuang2019single}.
The structure of 
DarkNet-RI
is illustrated in 
Fig.~\ref{fig:model},
with the details of each block listed in
Table~\ref{table:conv_detail}.
The network follows an encoder-decoder
structure, 
utilizing DarkNet-53 as the
encoder comprising convolutional and residual
blocks. The width and height of the
activation maps decrease during encoding,
to facilitate the scale invariant pyramid 
illustrated in~Fig.~\ref{fig:pyramid}.
The decoder subsequently upsamples
these activation maps
to generate pixel-wise labels at different scales.

The network uses a total of 23 residual blocks as in~\cite{redmon2018yolov3},
each containing 1$\times$1 and 3$\times$3 convolutional filters,
and 40 convolutional layers.
It uses Leaky ReLU as the activation function in each convolutional layer.
Four skip connections feed forward the encoder activations from
individual residual blocks,
performing inception-like~\cite{szegedy2015going} channel concatenation
with their respective same-dimensional
upsampled decoder activations.
Five decoder embeddings
are flattened and passed to fully connected layers
and used in the loss function at training.

\subsubsection{Pyramid Representation Learning Layer}
\begin{figure}[!t]
\begin{center}
\includegraphics[width=0.9\linewidth]{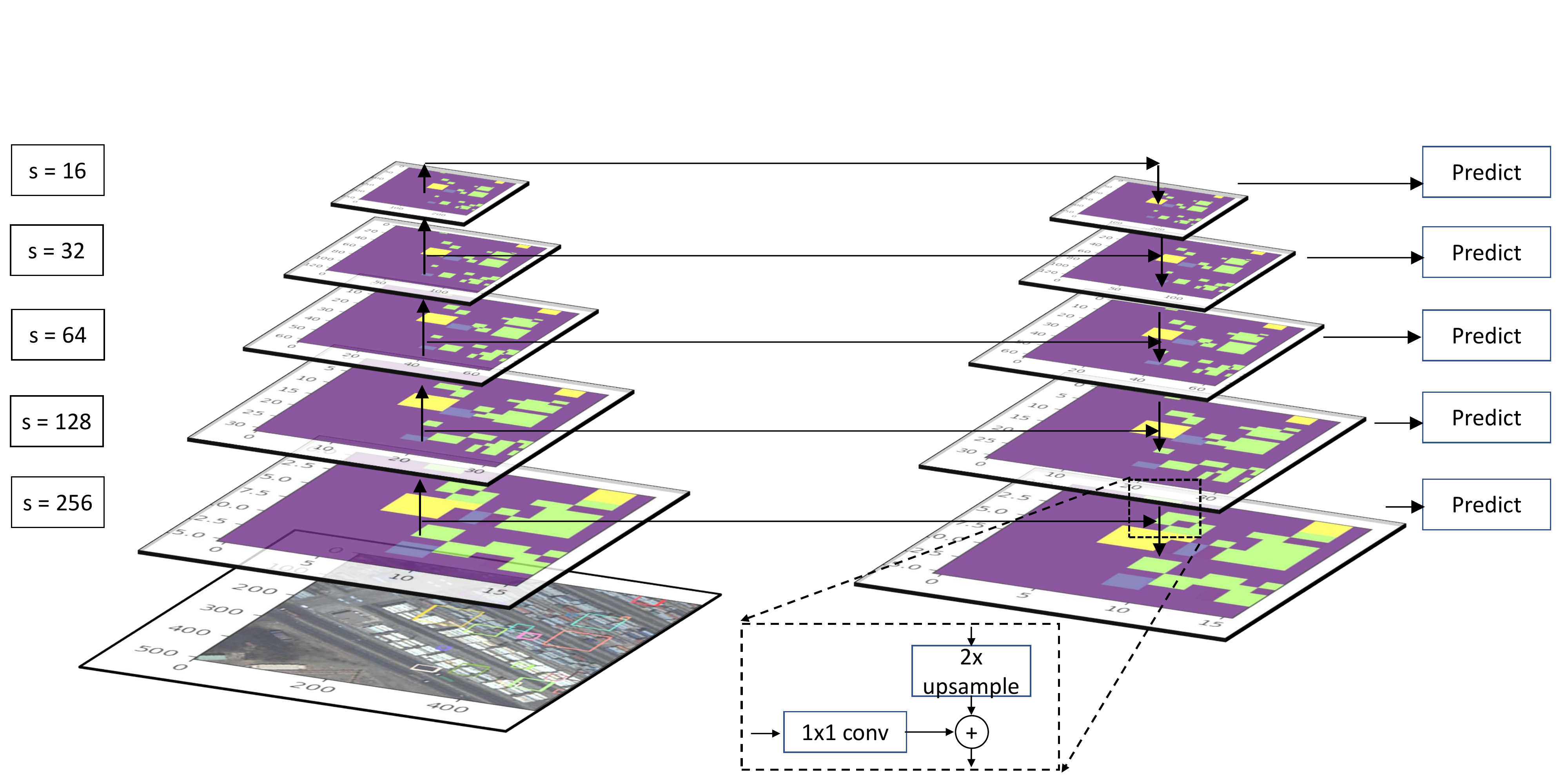}
\end{center}
   \caption{Multi-scale pyramidal feature map with CNN features at 5 layers.}
\label{fig:pyramid}
\end{figure}

Inspired by the Feature Pyramid Network (FPN)~\cite{lin2017feature}, we generate high-level semantic feature maps at different scales, which facilitates the detection of objects of different sizes. 
As illustrated in Fig.~\ref{fig:pyramid}, our network takes an arbitrary-sized image and convolves it with filters by strides of 2 to generate robust multiscale embeddings which are downsized by factors of 2, 4, 8, 16, and 32. 
Predictions are therefore achieved at 5 different scales corresponding to these 5 convolutional embeddings. 
Specifically, each embedding is upsampled by a factor of 2 and added to the corresponding convolutional embeddings from the encoder to generate the next finer feature map. So for example, as seen in Fig.~\ref{fig:model} the embedding output from Res1 is concatenated channel-wise with the output from layer Up4 and fed into ConvSet5.

\begin{figure}[!t]
\begin{center}
   \includegraphics[width=0.95\linewidth]{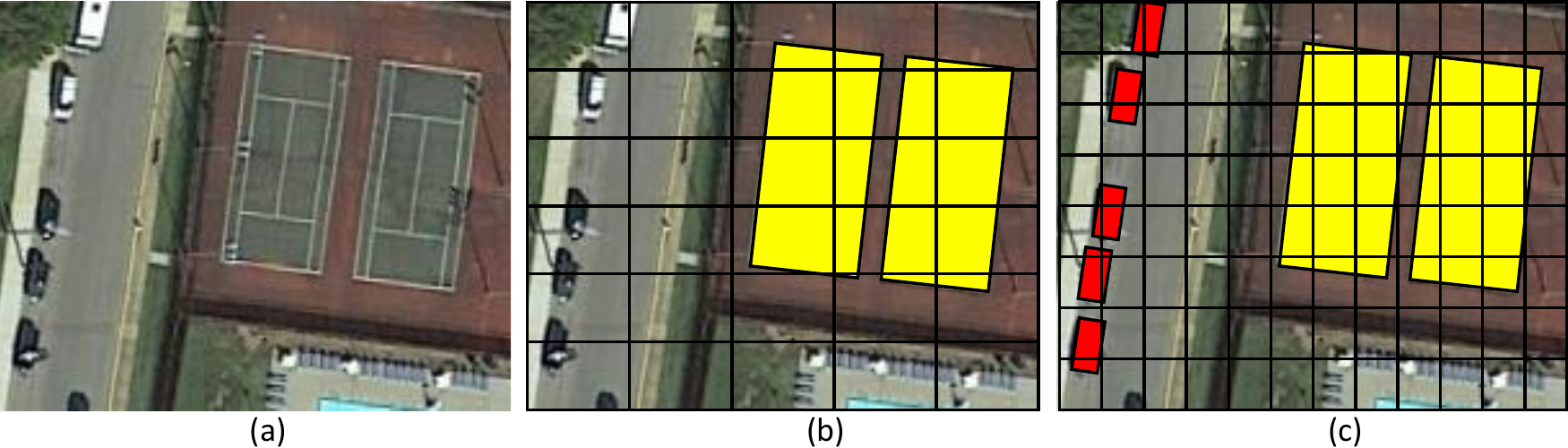}
\end{center}
   \caption{Multi-scale \emph{on-off} strategy on a sample image; a) ground-truth, b) objects which are smaller than the cell become 
   \emph{`off'}, c) objects (shown in red) become \emph{`on'} in a smaller cell.}
\label{fig:on-off}
\end{figure}

Our model takes advantage of the existing multiscale approaches~\cite{deng2018multi,li2018multiscale,guo2018geospatial} and designs a more robust representation for remote sensing imagery where images usually contain tiny objects as well as larger objects. It is capable of being used to detect more geospatial objects with different scales and shapes. More importantly, we specialize the layers in representation learning. Since the larger objects can be detected with less detail, the earlier layers are smoothed for these objects by easing their responsibility for smaller objects representation. 
This is achieved in the loss computation by making the ground truth objects \emph{on-off} based on their sizes for each layer. This is shown in Fig.~\ref{fig:on-off}. If the object is smaller than a grid cell at a given scale and therefore lies entirely within this cell, it will be \emph{off} in the corresponding feature map at that scale. Any finer scales that do not completely encompass the object within a single cell will then help in the feature representation, and the object will be \emph{on} in their feature maps. More specifically, if the size of a given object is smaller than $s/8$, then the feature map at scale $s$ will be \emph{off} for this object, and the next finer maps are therefore responsible for its feature representation. 
This might happen when there are several tiny objects in a dense area. When an object is larger than the grid cell, however, the predictions will be similar for the neighboring cells that contain parts of the object. In this case, each part participates in the final detection. Intuitively, each part of a certain object is considered at multiple scales in the corresponding feature maps. 

\subsubsection{In-Plane Rotation-Invariance}
There are two approaches to impart rotation-invariance, the first of which is data augmentation~\cite{van2018you,tayara2018object,redmon2016you}. In~\cite{van2018you} for instance, training images were rotated about the unit circle to improve the classifier robustness against rotation variations. Tayara and Chong~\cite{tayara2018object} used random rotations, vertical and horizontal flipping, and other transformations to generate artificially altered images of data samples within the dataset. The second approach to handle rotation-invariance is to modify the network structure through network design or regularization. An example is~\cite{li2017rotation,cheng2016rifd}, who leveraged rotation-invariant features into the CNN network. Another example is Li~\textit{et al.}~\cite{li2017rotation}, who added multiangle region proposals, in addition to the common multiscale and multiaspect-ratio ones, into their region proposal network (RPN). Cheng~\textit{et al.}~\cite{cheng2018learning} incorporated and learned a rotation-invariant layer into a CNN architecture.

As opposed to these methods, we do not add a new rotation-invariant layer into the network and instead we learn the rotation-invariant features by imposing a constraint in a regularization term only. Our method therefore does not add extra parameters to the network, which would increase the computation time and make the training suffer from overfitting.
Inspired by~\cite{cheng2018learning}, we optimize a new objective function to constrain the generalization of the training samples against rotation at each scale. It enforces the CNN features to be similar before and after applying rotations. We also first augment the input-output pairs by applying a random set of different rotations in the range of $360^\circ$ to the samples in the dataset, thereby adding new training examples to the dataset. 

Given the initial training samples $X\!=\!\{x_1,\dots,x_n\}$ \emph{before rotation} and applying $r$ in-plane rotations with angles drawn from the set
$\mathrm{A}\!=\!\{1^\circ,\dots,360^\circ\}$, 
the regularization constraint is formulated as:
\begin{equation}
\label{eq:rotation}
 G(X,X^R)=\frac{1}{r n}\sum_{\alpha \in \mathrm{A}} \sum_{x_i  \in X}{\| f(x_i)-f(R_\alpha(x_i)) \|_2^2}
\end{equation}
where
$R_\alpha(x_i)$
denotes an in-plane rotation of sample $x_i$ by angle
$\alpha$,
$f(x_i)$ and $f(R_\alpha(x_i))$ denote the extracted features of $x_i$
and $R_\alpha(x_i)$,
and $X^R$ denotes the \emph{after rotation} set for the rotated samples.
Here, $f(x_i)$ denotes the network outputs obtained from all 5 fully connected layers $\{FC_1,...FC_5\}$.
Apparently, the generated features for each sample are rotation-invariant if the output of this equation is small enough. These features are fed into the model at each scale for a robust feature representation.

\begin{figure}[!t]
\begin{center}
 \includegraphics[width=1\linewidth]{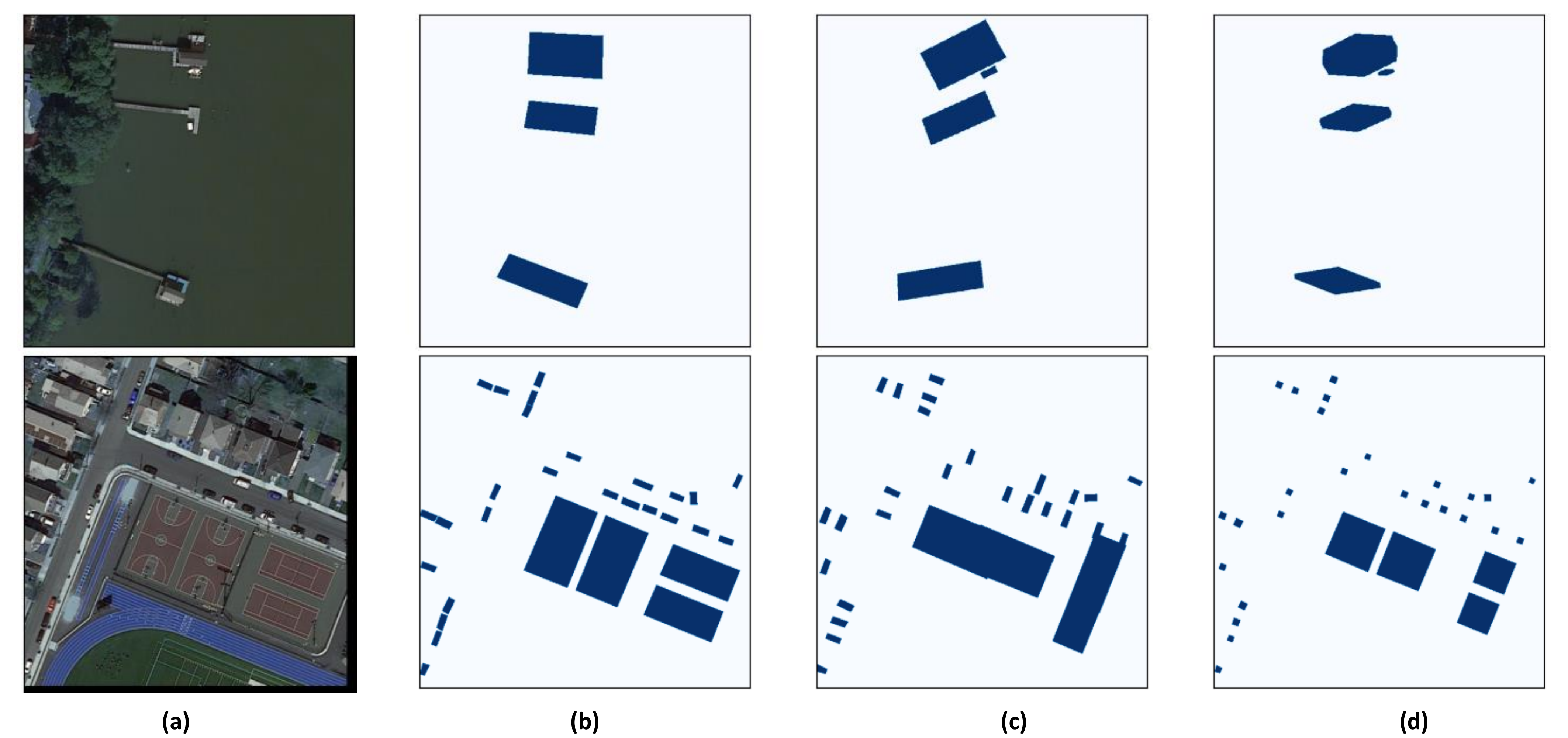}
\end{center}
   \caption{The area of overlap between the ground truth and rotated bounding boxes, within which the features are learned. Extracted features are then used in the rotation-invariant layer. (a) training images (b) bounding box, (c) $\alpha = 30^o$ (top), $\alpha=90^o$ (bottom), (d) area of overlap, over which rotation-invariant features are learned. }
\label{fig:rotation}
\end{figure}

To obtain the rotation samples, all objects are rotated around their centers. For each object, the intersection of the ground truth box and rotated box is then computed. Since our computations are pixel-wise, the regularization constraint only works on this intersection area and enforces all extracted features $f(x_i)$ of the object to be similar to their corresponding rotated features $f(R_\alpha(x_i))$ in their overlapping area. This is illustrated in Fig.~\ref{fig:rotation} where all objects inside the images in the first and second rows are respectively rotated by $ 30^\circ$ and $ 90^\circ$. It must be noted that although corner sides of the objects are lost in some rotation angles (they are not in the overlapping area), the entire shape of the bounding box is preserved when other constraints are considered between predictions and ground-truths in the training time. This is explained in the following section.

\subsubsection{Training}
The optimal parameters of the model are learned via training on the high-resolution images. For training our model, we use a multi-term loss function for the scale $s$ as follows: 
\begin{equation}\label{loss}
\ell^s  = \ell_{conf} + \ell_{class} + \ell_{rotation}\\
\end{equation}
\noindent where $\ell_{conf}$ denotes the object confidence score, $\ell_{class}$ is the classification loss and $\ell_{rotation}$ denotes the rotation errors computed using Eq.~\ref{eq:rotation}. 

The confidence score is computed for every cell location and shows if the corresponding location includes any objects. This is based on the original object before rotation, and therefore, the object shape is preserved. We use a softmax loss for the confidence score. 

The classification loss  $\ell_{class}$ is a cross entropy loss which is used to compute the loss between the predicted class for each object location and the ground truth object class for that location to penalize classification errors. It is defined as:
\begin{equation}
\ell_{class}(c,\hat{c})=-\sum^C_{i=1} c_i log(\hat{c_i})
\end{equation}
\noindent where $C$ denotes the total number of classes, and $c$ and $\hat{c}$ represent the ground-truth and predicted labels, respectively. Notably, $c$ is the one-hot ground truth vector and $\hat{c}$ is the predicted probability vector after applying softmax activation. This loss enforces that each object be associated with a single class with the maximum probability, such that increasing the likelihood for one class will necessarily decrease it for  other classes. The losses are computed for each scale and are summed for the final network outputs as $\ell = \sum\limits_s \ell^s$.

\subsection{Oriented Box Determination}
As opposed to the existing approaches, our method is a pure classification solution. Other methods use regression-based bounding box detection which is prone to error specifically for oriented bounding boxes, in which the number of outputs are greater (i.e., 8 values, representing the coordinates of the 4 corners). The proposed method generates object proposals for each object at different scales from coarser to finer scales through $FC_1$ to $FC_5$. It particularly predicts class labels along with confidence scores for image locations which are uniform-sized cells in the grids at multiple scales. Notably, the bounding box location and dimension are not explicitly predicted. We infer the bounding boxes only at inference time by finding the minimum surrounding box for every set of same predicted class labels which form a connected neighborhood. This is achieved by a set of morphological operations. 

At the inference stage, the network returns 
a set of class label matrices and confidence score matrices
which are then used as inputs to the oriented bounding box determination module. As network outputs are pixel-wise, these matrices are image-sized and they are created per scale level.

The class label matrix includes pixel labels similar to the outputs in the semantic segmentation task~\cite{girshick2014rich, kemker2018algorithms}. A set of morphological operations~\cite{duan2019large} (e.g., \emph{denoising, erosion, opening}, and \emph{closing}) are then applied on to smooth and remove noise from these matrices.
The refined matrices are therefore obtained from 5 class label matrices which correspond to 5 scale levels. At a specific level, we treat the corresponding matrix as a binary matrix and find the contours for each set of the same predicted class labels. This is simply achieved by using the Suzuki algorithm~\cite{suzuki1985topological} for contour detection. From each closed region, a bounding box with the minimum area is then extracted. To do so, the convex hull of the contour point set is first found~\cite{sklansky1982finding}, and based on the observation that a side of a minimum-area enclosing box must be collinear with a side of the convex polygon, the smallest enclosing box is obtained using the rotating calipers algorithm~\cite{toussaint1983solving}. The algorithm returns a 2D box structure as $(x,y, w, h, \alpha)$, where $(x,y)$ denotes the box center, $(w,h)$ respectively denote the width and height of the box, and $\alpha$ represents the angle of rotation. It must be noted that these algorithms are implemented in OpenCV library~\cite{bradski2008learning}. 

For each box, we finally extract the box corner points which inherently include the box rotation. Object labels are also assigned to these boxes from their corresponding regions in the class label matrix. As a result, all image pixels are labeled as \emph{background} or as an object class. The confidence score also shows the probability of every pixel to be an object label. Furthermore, the minimum bounding box is detected for each region regardless of the orientation of the box. A set of prediction vectors are finally achieved per scale level,
with each predicted oriented bounding box represented as
$(x_1,y_1,x_2,y_2,x_3,y_3,x_4,y_4,c)$,
or optionally reduced to $(x_1,y_1,x_2,y_2,c)$
if the bounded box is axis aligned.

Our method is capable of detecting objects with different sizes. In particular, pixels are mapped to grid cell locations at different scales and ground-truth labels are assigned to these cells. The detections are therefore pixel-wise per scale level.

\subsection{Box Refinement}
The predictions are finally used as inputs to the NMS algorithm~\cite{neubeck2006efficient} to remove overlapping boxes based on threshold values of their Intersection over Union (IoU) metric. The NMS algorithm is commonly used in object detection approaches to eliminate multiple false positive bounding boxes by filtering against a single threshold value. This value must be non-zero, to admit some degree of overlap between adjacent bounding boxes. 
Setting the threshold too small, however, will result in false negatives, as some true small objects will be removed.

To address this issue, we propose suppressing  bounding boxes by considering their IoU scores at multiple scale levels, using a distinct threshold value for each scale. Since there are 5 scale levels, 5 threshold values $\{\theta_1,...,\theta_5\}$ are defined to suppress overlapped boxes in each corresponding scale. For each scale $i$, we calculate an IoU for each bounding box with all other boxes, and eliminate those boxes which have a higher IoU than the scale threshold $\theta_i$. In this way, adjacent boxes are therefore compared only at the same scale level. Moreover, we use an average confidence score computed using all individual cells inside each detected box. Our averaged confidence score is impacted by each cell's confidence score. This is different than the existing approaches which rely on only a single IoU threshold value for each box.  This score can better downgrade the low-quality bounding boxes. Furthermore, a detected bounding box is considered to be low quality if its average confidence score is low.

\section{Experiments}
Two challenging and large-scale datasets, xView~\cite{lam2018xview} and DOTA~\cite{xia2018dota}, were selected to evaluate the proposed method for remote sensing imagery. These datasets are described in the following subsections, along with the  results of a set of experiments comparing the performance of the proposed approach to other leading methods on a set of challenging object recognition tasks, as described in detail below.

\subsection{Datasets and Protocols}

\noindent \textbf{xView Dataset:} This dataset covers a total area of 1,400 $km^2$ at 0.3 meter ground sample distance (GSD: the distance between pixel centers as measured on the ground),
and includes approximately 1 million labeled ground objects from 60 categories. It contains 846 high-resolution annotated images of varying sizes ranging from 2564$\times$2576 to 3187$\times$4994 pixels.

\noindent \textbf{DOTA Dataset:} This dataset consists of 2806 images with objects from 15 different categories. DOTA images range from 800$\times$800 to 4000$\times$4000 pixels and are captured using different sensors. There are 188,282 object instances labeled by oriented bounding boxes. The DOTA dataset is used to evaluate remote sensing object detectors in two tasks, i.e., horizontal bounding box detection (HBB), and oriented bounding box detection (OBB). We use this dataset for the OBB task. 
The short names for the 15 categories are given as: Plane (PL), Baseball diamond (BD), Bridge (BR), Ground field track (GTF), Small vehicle (SV), Large vehicle (LV), Ship (SH), Tennis court (TC), Basketball court (BC), Storage tank (ST), Soccer-ball field (SBF), Roundabout (RA), Harbor (HA), Swimming pool (SP), and Helicopter (HC). 

\noindent \textbf{Comparison:} The mean average precision (mAP) was used as the standard criteria to compare our model with other current state-of-the-art methods. We selected both general-purpose object detection methods, as well as models specifically developed for remotely sensed images. To this end, SSD~\cite{liu2016ssd}, YOLOv3~\cite{redmon2018yolov3}, YOLT~\cite{van2018you}, and Reduced Focal loss (RFL)~\cite{sergievskiy2019reduced} were selected as comparison baselines for HBB task on the xView dataset. For the OBB task, we compared our method with R-DFPN~\cite{yang2018automatic}, RRPN~\cite{ma2018arbitrary}, ICN~\cite{azimi2018towards}, SCRDet~\cite{yang2019scrdet}, RoI Transformer~\cite{ding2019learning}, SARD~\cite{wang2019sard}, O$^2$-DNet~\cite{wei2019oriented}, CAD-Net~\cite{zhang2019cad}, and P-RSDet~\cite{zhou2020objects}. The backbone of these methods is given in Table~\ref{table:backbones}.

\begin{table}[h]
\caption{Network backbones of the comparison methods.}
\label{table:backbones} 
\centering
\small
\begin{tabular}{p{3cm} P{4cm}}
  \hline
  Method & Backbone \\ 
    \hline\hline
  R-DFPN~\cite{yang2018automatic} & ResNet-101 \\ RRPN~\cite{ma2018arbitrary} & VGG-16~\cite{simonyan2014very} \\ ICN~\cite{azimi2018towards} & ResNet-101 \\ SCRDet~\cite{yang2019scrdet} & ResNet-101 \\ 
  RoI Transformer~\cite{ding2019learning} & ResNet-101\\ 
  SARD~\cite{wang2019sard} & ResNet-101 \\ O$^2$-DNet~\cite{wei2019oriented} & 104-Hourgalss + CornerNet \\ CAD-Net~\cite{zhang2019cad} & ResNet-101 \\ P-RSDet~\cite{zhou2020objects} & ResNet-101 \\

  \hline
\end{tabular}
\end{table}

\begin{figure}[!t]
\begin{center}
 \includegraphics[width=1.1\linewidth]{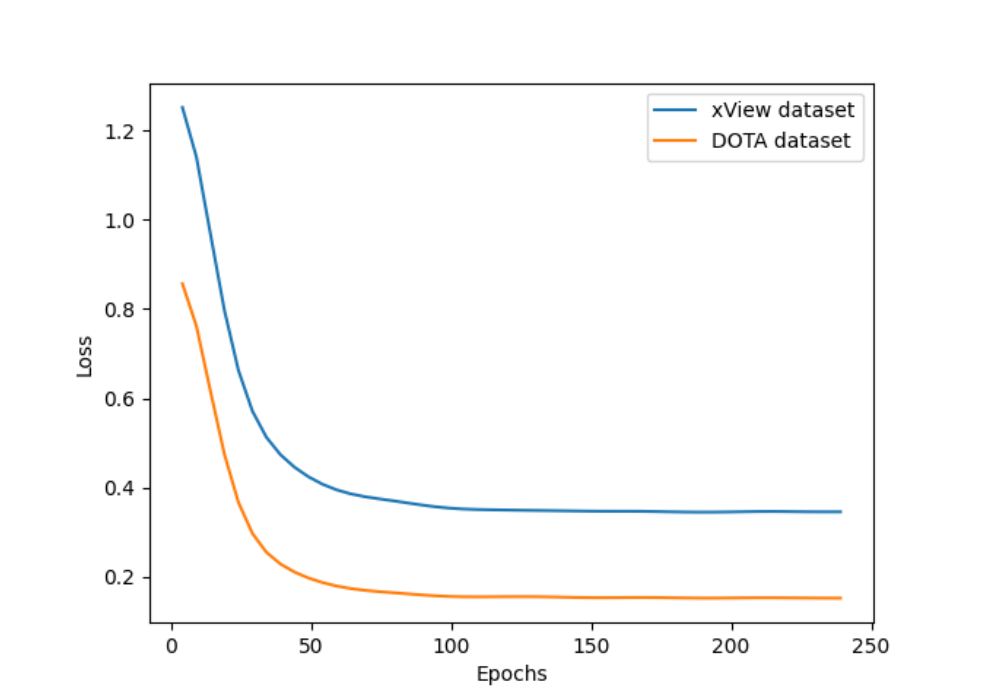}
\end{center}
   \caption{Training loss of the proposed model on both xView and DOTA datasets.}
\label{fig:training_curves}
\end{figure}

\noindent \textbf{Implementation:} The proposed method accepts images of 512$\times$512 pixels as input. We tiled the high-resolution images from both datasets into 512$\times$512 pixels with an overlapping area of 10 pixels,  using zero padding to keep the aspect ratios of non-square images after splitting the original images. The hyper-parameters were set as initial learning rate = 0.01, momentum = 0.9, and weight decay = 0.0005. The learning rate was reduced by an order of magnitude (multiplied by a factor of 0.1) for each 50000 iterations. The experiments were executed on a single Nvidia Titan RTX GPU and the batch size was set to 4. Fig.~\ref{fig:training_curves} shows  the  model’s  loss  during  training on both datasets,  which  allows  the network  to  dynamically  be  tuned. We trained our network on both xView and DOTA datasets for 240 epochs. Also, it must be noted that our method achieved a reasonable inference time on these datasets. It particularly runs on average at 60 and 10 ms per image for images of the xView and DOTA datasets, respectively. This time includes algorithms in all modules.

\begin{table*}[h]
\caption{Performance (mAP) on xView dataset, 19 small object classes with high between-class similarity.}
\label{table:tinyObjects} 
\centering
\small
\begin{tabular}{p{2cm} P{2cm} P{2cm} P{2cm} P{2cm} P{2cm}}
  \hline
  Object classes & SSD \cite{liu2016ssd} & YOLOv3 \cite{redmon2018yolov3} & YOLT \cite{van2018you} & RFL \cite{sergievskiy2019reduced} & \textbf{Ours} \\
  \hline\hline
  Bus & 0.3773 & 0.3572	& 0.4321	& 0.4475	& \textbf{0.4685}
\\
Cargo truck  & 0.0972	& 0.0892	& 0.1121	& 0.1147	& \textbf{0.1345}
\\
Cement mixer & 0.1441	& 0.1432	& 0.229	& 0.238	& \textbf{0.2645}
\\
Crane truck  & 0.0838	& 0.1019	& 0.121	& 0.1218	& \textbf{0.126}
\\
Dump truck  & 0.2275	& 0.228	& 0.2518	& 0.261	& \textbf{0.2645}
\\
Excavator & 0.4691	& 0.4627	& 0.4825	& 0.4826	& \textbf{0.4945}
\\
FL/B & 0.322	& 0.3382	& 0.359	& 0.3682	& \textbf{0.377}
\\
Ground grader  & 0.191	& 0.1928	& 0.2534	& 0.2674	& \textbf{0.298}
\\
Motorboat  & 0.2488	& 0.2227	& 0.2745	& 0.2639	& \textbf{0.2965}
\\
PV & 0.5569	& 0.425	& 0.5625	& 0.553	& \textbf{0.5955}
\\
Pickup truck & 0.0078	& 0.0126	& 0.1063	& 0.1296	& \textbf{0.136}
\\
Scraper tractor  & 0.1109	& 0.1092	& 0.1522	& 0.1728	& \textbf{0.208}
\\
SC  & 0.3835	& 0.358	& 0.4276	& 0.4184	& \textbf{0.465}
\\
Small car & 0.4083	& 0.3328	& 0.4922	& 0.4729	& \textbf{0.544}
 \\
Trailer & 0.2151	& 0.228	& 0.231	& 0.259	& \textbf{0.2625}
\\
Truck & 0.0469	& 0.071	& 0.1022	& 0.125	& \textbf{0.1675}
\\
Truck tractor & 0.2129	& 0.218	& 0.2361	& 0.2395	& \textbf{0.2435}
\\
TT/FT & 0.1261	& 0.1178	& 0.152	& 0.1574	& \textbf{0.178}
\\
Utility truck  & 0.2846	& 0.2557	& 0.2912	& 0.2845	& \textbf{0.299}
\\
\hline
\textbf{Overall}  & 0.2376	& 0.2244	& 0.2773	& 0.2830	& \textbf{0.3065}
\\
  \hline
\end{tabular}
\end{table*}

\subsection{xView Dataset Experiment}
The tiling was performed for both training and test images. Each cropped training set image was augmented with scaling, vertical and horizontal flipping, shearing, and translation. In order to create the \emph {before rotation} and \emph{after rotation} sets, each augmented cropped image was also rotated with 5 angles randomly selected from $\{30,60,90,120,150,180,210,240,270,300,330\}$ degrees, and a total number of 123,750 images was finally obtained. A ratio of 80\% of these were used for training and validation of the model and the remaining 20\% were employed for testing.

We report the overall performance on the whole test set, as well as the separate performance on the detection of tiny objects, which were categorized into 19 classes as: \emph{Bus, Cargo truck, Cement mixer, Crane truck, Dump truck, Excavator, Front loader/Bulldozer, Ground grader, Motorboat, Passenger vehicle, Pickup truck, Scraper tractor, Shipping container, Small car, Trailer, Truck, Truck tractor, Truck tractor w/ flatbed trailer, and Utility truck.} 

To evaluate the performance for tiny object detection, we performed an experiment with only these small object classes. Table~\ref{table:tinyObjects} reports the mAP results on these 19 challenging classes. (\emph{Front loader/Bulldozer, Passenger vehicle, Truck tractor w/ flatbed trailer, and Shipping container} are shown in the table with their abbreviations as FL/B, PV, TT/FT, and SC, respectively.) The object instances of all 19 classes are not only small and dense, but the number of instances of each class in the dataset are imbalanced, with very high between-class similarity. Detecting these tiny objects requires high resolution images and a dense grid cell. 

The proposed method achieved the highest accuracy for each of the 19 classes, with an overall mAP of 0.3065, which is significantly higher than that of SSD at 0.2376, YOLOv3 at 0.2244, YOLT at 0.2773, or RFL at 0.2830. Notably, the proposed method was able to detect even small cars with a mAP of 0.544. 

The proposed method combines the advantages of multi-scale features along with pixel-wise bounding box prediction. Since the proposed method uses a dense feature map of 256$\times$256 at the finest scale, it is able to detect tiny objects as small as 2$\times$2 pixels. This means objects as small as $0.6$ \emph{meters}$^2$ can be detected with a GSD of $0.3$ meters. Each grid cell is therefore small enough to encompass the smallest object (i.e., \emph{Small car}) as a single prediction unit. The best detected classes are: \emph{Passenger vehicle, Small car, Excavator, Bus, Shipping container}. It must be noted that the instance counts in these classes are relatively high compared with  those of other classes. Therefore both average pixel area and instance count can be seen to benefit effective detection.

\begin{figure*}[!t]
\begin{center}
 \includegraphics[width=0.9\linewidth]{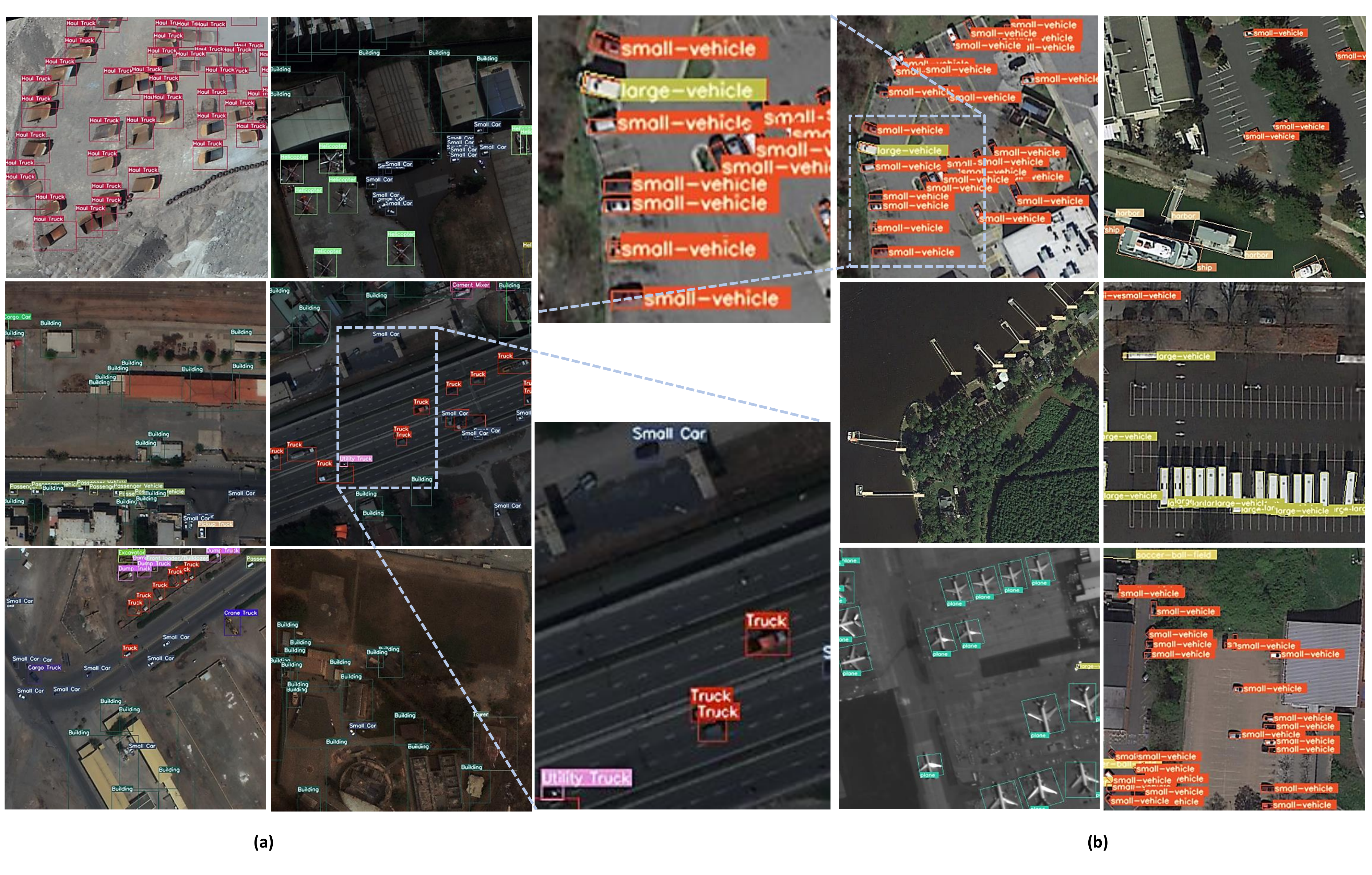}
\end{center}
   \caption{Sample detection results from xView dataset (a) and DOTA dataset (b). Most of the tiny objects such as small cars are correctly detected.}
\label{fig:examples}
\end{figure*}

The overall performance is compared with the state-of-the-art methods in Table~\ref{table:xview_performance}. YOLT obtained the second best results with a mAP of 0.49. It uses an ensemble of multiple detectors at multiple scales, and employs a denser grid compared to YOLOv3. The general-purpose methods do not take  advantage of high resolution features which are essential to detect tiny objects. The dense grid is also crucial for detecting tiny objects in high density areas. The spatial constraints in these methods limit the number of nearby objects that they can predict. In anchor-based methods, many anchor boxes are used at each location over the feature maps with the additional IoU computation (\emph{e.g} 9 anchor boxes per location in YOLOv3), while most of them do not specify any object. Our method however can detect bounding boxes without the need for IoU computation and the hassle of anchor design. The performance boost is evident with the proposed method which achieved the best mAP performance of 0.5218. Some detected objects in a dense area are depicted in Fig.~\ref{fig:examples}(a).

\begin{table}[h]
\caption{Performance comparison results on xView dataset, all 60 object classes.}
\label{table:xview_performance}
\centering
\begin{tabular}{p{4cm} P{1cm}}
\hline
Method & mAP \\
\hline\hline
SSD \cite{liu2016ssd} & 0.3200 \\
YOLOv3 \cite{redmon2018yolov3} & 0.2420 \\
YOLT \cite{van2018you} & 0.4900\\
RFL \cite{sergievskiy2019reduced} & 0.3174\\
\textbf{Ours} & \textbf{0.5315}\\
\hline
\end{tabular}
\end{table}

\subsection{DOTA Dataset Experiment}
We used the same settings for our experiments on the DOTA dataset. For splitting the images, we used the provided development kit, with the splitting rate set to 2. By removing the non-labeled image parts, a total of 49,601 images were generated for training. We selected another splitting rate as 1 and generated 14,348 images, from which our rotated set was created containing 71,740 images. 

We evaluated the performance of our object detection approach on oriented bounding box (OBB) detection task of the DOTA dataset. The overall performance on all 15 categories is reported in Table~\ref{table:dota_performance}. It can be seen that our method outperforms all other methods evaluated on this dataset. It specifically achieves 75.5 mAP which is 2.55\% higher than the second best method, which is SARD~\cite{wang2019sard} with 72.95 mAP. We argue that feature fusion can effectively impact the detection results. The three approaches (i.e., O$^2$-DNet, SCRDet, and RoI-Transformer) for which feature fusion
is a central 
design element can detect small objects better than the other methods which rely less on feature future. SCRDet also expands the receptive field, which leads to better detection for larger objects. 

These methods however are all sensitive to the angle and IoU in the bounding boxes with large aspect ratios. In particular, SCRDet suppresses the noise influence and highlights the object information through an attention network. It however regresses the bounding box angle in addition to the bounding box center, width and height, which significantly increases ambiguity. Likewise, the other methods regress the additional values (angle, height, or 4 values corresponding to corner points coordinates) for bounding box detection. Our method however casts the overall task of object detection to a classification problem which doesn't add the ambiguity or complexity to deal with oriented boxes. The results show that our method ranks first among the other methods in oriented bounding box detection in 8 categories. 

Based on the results on both axis-aliened and oriented box detection tasks, we can draw the conclusion that our method is a better object detector for remote sensing applications. We believe that this is due to the multi-scale and rotation-invariant feature representation learning, and more importantly, casting the sensitive regression task to a pure classification problem.

\begin{table*}[h]
\caption{Performance comparison results of OBB task on  DOTA dataset. The best result in each category is highlighted in bold.}
\setlength
\tabcolsep{5pt}
\label{table:dota_performance}
\scriptsize 
\centering
\begin{tabular}{lllllllllllllllll}
\hline
Method & PL & BD & BR & GTF & SV & LV & SH & TC & BC & ST & SBF & RA & HA & SP & HC & mAP \\
\hline\hline
R-DFPN~\cite{yang2018automatic} & 80.92 & 65.82 & 33.77 & 58.94 & 55.77 & 50.94 & 54.78 & 90.33 & 66.34 & 68.66 & 48.73 & 51.76 & 55.10 & 51.32 & 35.88 & 57.94 \\
RRPN~\cite{ma2018arbitrary} & 88.52 & 71.20 & 31.66 & 59.30 & 51.85 & 56.19 & 57.25 & 90.81 & 72.84 & 67.38 & 56.69 & 52.84 & 53.08 & 51.94 & 53.58 & 61.01 \\
ICN~\cite{azimi2018towards} & 81.40 & 74.30 & 47.70 & 70.30 & 64.90 & 67.80 & 70.00 & 90.80 & 79.10 & 78.20 & 53.60 & 62.90 & 67.00 & 64.20 & 50.20 & 68.20 \\
SCRDet~\cite{yang2019scrdet} & \textbf{89.98} & 80.65 & 52.09 & 68.36 & 68.36 & 60.32 & 72.41 & 90.85 & \textbf{87.94} & 86.86 & \textbf{65.02} & 66.68 & 66.25 & 68.24 & 65.21 & 72.61 \\
RoI Trans.~\cite{ding2019learning} & 88.64 & 78.52 & 43.44 & 75.92 & 68.81 & 73.68 & 83.59 & 90.74 & 77.27 & 81.46 & 58.39 & 53.54 & 62.83 & 58.93 & 47.67 & 69.56 \\
SARD~\cite{wang2019sard} & 89.93 & \textbf{84.11} & \textbf{54.19} & 72.04 & 68.41 & 61.18 & 66.00 & 90.82 & 87.79 & 86.59 & 65.65 & 64.04 & 66.68 & 68.84 & \textbf{68.03} & 72.95 \\

O$^2$-DNet~\cite{wei2019oriented} & 89.31 & 82.14 & 47.33 & 61.21 & 71.32 & 74.03 & 78.62 & 90.76 & 82.23 & 81.36 & 60.93 & 60.17 & 58.21 & 66.98 & 61.03 & 71.04 \\

CAD-Net~\cite{zhang2019cad} & 87.8 & 82.4 & 49.4 & 73.5 & 71.1 & 63.5 & 76.7 & \textbf{90.9} & 79.2 & 73.3 & 48.4 & 60.9 & 62.0 & 67.0 & 62.2 & 69.9 \\

P-RSDet~\cite{zhou2020objects} & 89.02 & 73.65 & 47.33 & 72.03 & 70.58 & 73.71 & 72.76 & 90.82 & 80.12 & 81.32 & 59.45 & 57.87 & 60.79 & 65.21 & 52.59 & 69.82 \\

\textbf{Ours}  & 89.0	& 80.4	& 50.5	& \textbf{76.0}	& \textbf{78.2}	& \textbf{77.8}	& \textbf{87.4}	& 90.0	& 83.2	& \textbf{86.93}	& 63.3	& \textbf{68.7}	& \textbf{67.6}	& \textbf{70.8}	& 63.8 & \textbf{75.5}	\\
\hline
\end{tabular}
\end{table*}

\subsection{Ablation Study}
We performed other experiments to further investigate the efficiency of the proposed method. The effectiveness of our rotation invariant feature learning, multi-scale feature representation, and multi-scale semantic segmentation are specifically verified in the following subsections. 

\subsubsection{Rotation-Invariant Feature Learning}
To show the impact of the in-plane rotation-invariant feature learning, we designed three experiments on the xView dataset. All three tests
used the exact same two training and test image sets. In the first and second experiments, the rotation-invariant feature learning was removed from the network, by zeroing the rotation loss term in Eq.~\ref{loss} (i.e. 
$\ell_{rotation}=0$).
The rotated image augmentation 
was excluded from the training dataset for this first experiment, to evaluate the network's performance without any rotation augmentation. In the second experiment, 
while the rotation term of the
loss function remained zero,
some limited random rotations in the
range 0 to 10 degrees were added back to the training dataset for each cropped image, similar to the standard augmentation techniques used in existing methods. The third experiment added the full rotation
augmentation from 0 to 360 degrees
back to the training set, 
and included the rotation-invariant term  $\ell_{rotation}$
back to the loss function  Eq.~\ref{loss}. As indicated in Table~\ref{table:rotation_performance}, training on the augmented dataset outperforms the non-augmented results. Although augmenting the dataset with a limited number of image rotations can be useful, the extracted features for each object cannot comprehensively cover different rotation angles. This is why our method achieved slightly higher performance using the rotation-invariant feature representation. It also must be noted that the objects in the xView images are already captured in random orientations. 

\begin{table}[h]
\caption{Comparison results with/without rotation-invariant representation. Though rotation augmentation can enhance the performance, it is still not enough to cover the features of freely rotated objects in remotely sensed imagery.}
\label{table:rotation_performance}
\centering
\begin{tabular}{lll}
\hline
Method & Range & mAP \\
\hline\hline
Without rotation augmentation & - & 0.5018 \\
With rotation augmentation & (0-180) & 0.5185 \\
Using rotation-invariant & (0-90) & 0.5250\\
Using rotation-invariant & (0-180) & 0.5300\\
Using rotation-invariant & (0-360) & 0.5315\\
\hline
\end{tabular}
\end{table}

\subsubsection{Multi-Scale Feature Representation} 
\begin{figure}[!t]
\begin{center}
 \includegraphics[width=0.9\linewidth]{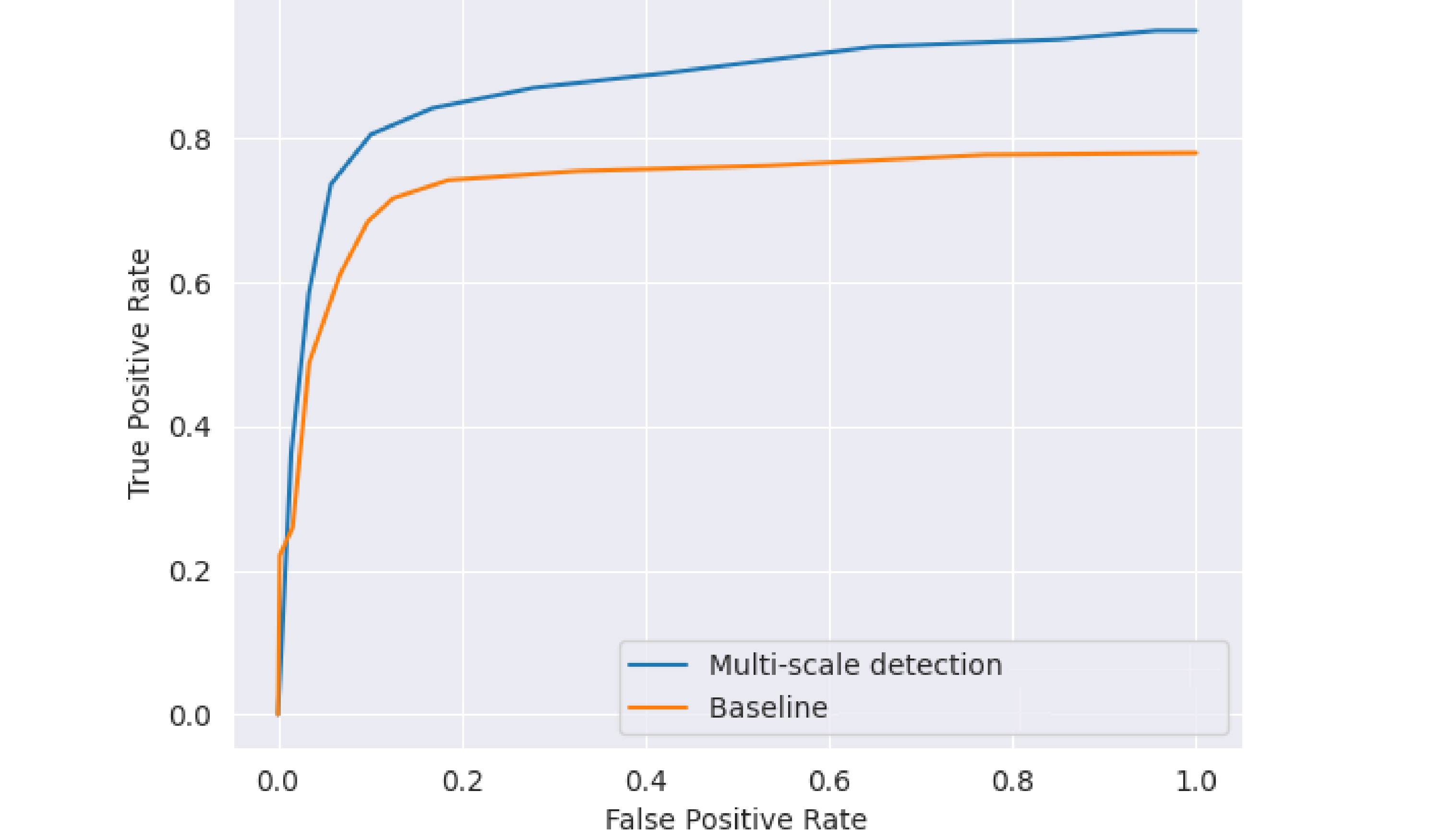}
\end{center}
   \caption{The ROC curve of the baseline method without multi-scale feature maps and our multi-scale feature representation approach on the DOTA dataset.}
\label{fig:roc_curves}
\end{figure}

We also designed an experiment on the DOTA dataset to investigate the impact of the multi-scale feature representation in our approach. We selected the baseline method by eliminating all intermediate fully convolutional layers (i.e. FC1 - FC4) in the decoder part of the network as inputs to the loss function.
It therefore included only one feature map of size 256$\times$256. 
The other layers and rotation-invariant feature representation however remained the same as our DarkNet-RI network. As depicted in Fig.~\ref{fig:roc_curves}, the baseline method dropped its performance substantially. It is clearly observed that some categories with larger-scale instances obtained lower performance when multi-scale feature maps are removed from our network. We also see a noticeable improvement in these categories using our approach. Nevertheless, some classes such as \emph{small-vehicle} and \emph{ship} do not benefit nearly as much as other classes from multi-scale feature representation.

\subsubsection{Multi-Scale Semantic Segmentation} The effectiveness of our multi-scale semantic segmentation for object detection was verified by conducting an experiment on the DOTA dataset. To directly compare the impact of our semantic segmentation method on the complete process, we replaced this module with other established and popular semantic segmentation networks. We selected UNet~\cite{ronneberger2015u}, SegNet~\cite{badrinarayanan2017segnet}, and SCAttNet~\cite{li2020scattnet} which have been previously used for remote sensing applications.

UNet consists of an encoder-decoder architecture. In the encoder part, it performs down-sampling by $3\times3$ convolutions, each followed by a ReLU and a $2\!\times\!2$ max-pooling operation with stride $2$. At each down-sampling step, the number of feature channels are doubled. In the decoder part, the feature map is up-sampled and the number of feature channels is halved by a $2\!\times\!2$ convolution. The feature map is then concatenated with the corresponding map from the encoder, and followed by two $3\!\times\!3$ convolutions, each followed by a ReLU. At the final layer, a $1\!\times\!1$ convolution is used to map each 64-component feature vector to the number of class labels. In total, the network includes 23 convolutional layers.

SegNet consists of a similar encoder-decoder architecture. The feature maps in the encoding stage are concatenated to up-sampled feature maps in the decoding part. The encoder network includes 13 convolutional layers, followed by a $2\times2$ max-pooling operation with stride $2$. Since the boundary delineation is vital for the segmentation task, the boundary information is stored in the encoder by memorizing the locations of the maximum feature values in each pooling window for each encoder feature map. In the decoder part, the feature map is up-sampled using the memorized max-pooling indices.

SCAttNet (semantic segmentation Network with Spatial and Channel Attention) consists of a ResNet50 backbone and an attention module. The attention module comprises a cascade of channel attention and spatial attention. Extracted features from the backbone network are fed into the channel attention module to refine the features in channels. The refined channel feature map is then used in the spatial attention module to refine the spatial axis. The semantic segmentation results are finally obtained by convolution and softmax operations. 

As our object detection method is modular and flexible, we replaced DarkNet-RI with the above-mentioned state-of-the-art semantic segmentation algorithms. Thus, the same process was performed on each of the segmentation results to obtain the final bounding boxes. It must be noted that pixels in the bounding boxes were used as ground-truths for all methods, whereas semantic segmentation algorithms are originally designed to learn the pixel features inside the object contours.

\begin{figure}[!t]
\begin{center}
 \includegraphics[width=0.9\linewidth]{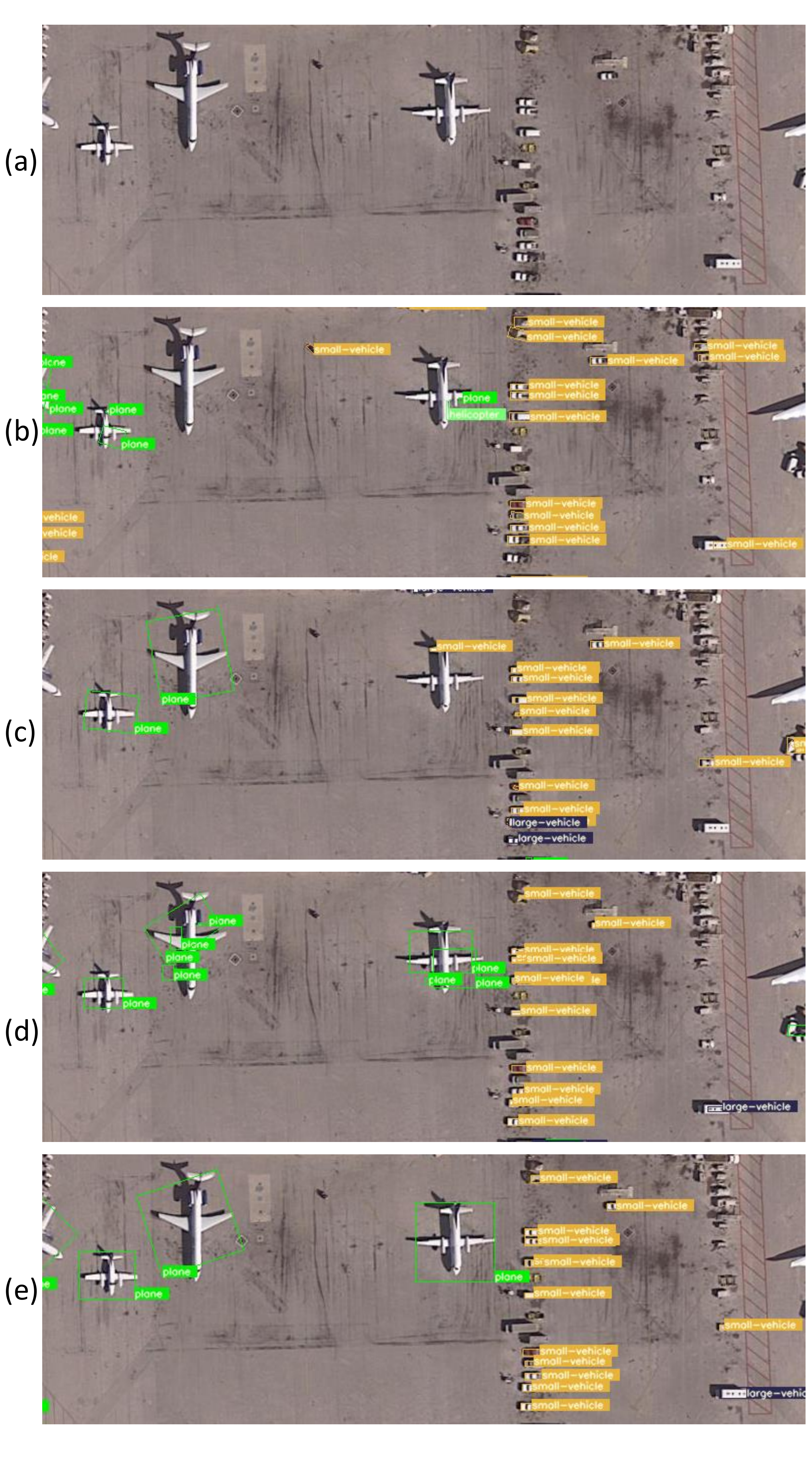}
\end{center}
   \caption{Object detection results using different semantic segmentation methods. From top to bottom, (a) shows the input image from DOTA dataset, and (b)-(e) respectively represent the detection results obtained by UNet~\cite{ronneberger2015u}, SegNet~\cite{badrinarayanan2017segnet}, SCAttNet~\cite{li2020scattnet}, and our DarkNet-RI.}
\label{fig:sem_seg}
\end{figure}

The final object detection results are shown in Table~\ref{table:ablation_semseg}. Our method demonstrates a mAP of $75.5\%$, which is significantly better that the other methods, the best of which was SCAttNet at $61.9\%$. One obvious reason is the impact of multi-scale detection which  benefits from our multi-scale feature representation. Our method uses the outputs of the semantic segmentation stage, which are class labels at 5 different scales. This allows us to detect objects in a wide range of scales, which naturally exist in remote sensing images. Another important reason is that our method provides more coherent labeling through merging the information of the up-sampling and down-sampling blocks. Although the other methods show relatively good results for small objects, they obtain inconsistent detections for the other objects. This is illustrated in Figure~\ref{fig:sem_seg}, where our method assigns correct class labels and bounding boxes to many objects of various sizes and orientations. This figure demonstrates that our method can detect objects as tiny as \emph{`small-vehicles'} while still correctly detecting other larger objects such as \emph{`large-vehicles'} and \emph{`planes'}.

\begin{table}[h]
\caption{Performance comparison results of OBB task using different semantic segmentation algorithms on  DOTA dataset.}
\label{table:ablation_semseg}
\centering
\begin{tabular}{lll}
\hline
Method & Architecture & mAP \\
\hline\hline

UNet~\cite{ronneberger2015u} & 23 convolutional layers & 55.4 \\
SegNet~\cite{badrinarayanan2017segnet} & Convolutional encoder-decoder & 60.4 \\
SCAttNet~\cite{li2020scattnet} & ResNet50 + spatial and channel  attention  & 61.9 \\
\textbf{Ours}  & \textbf{Multi-scale DarkNet} & \textbf{75.5}	\\
\hline
\end{tabular}
\end{table}

\begin{figure*}[!t]
\begin{center}
 \includegraphics[width=0.8\linewidth]{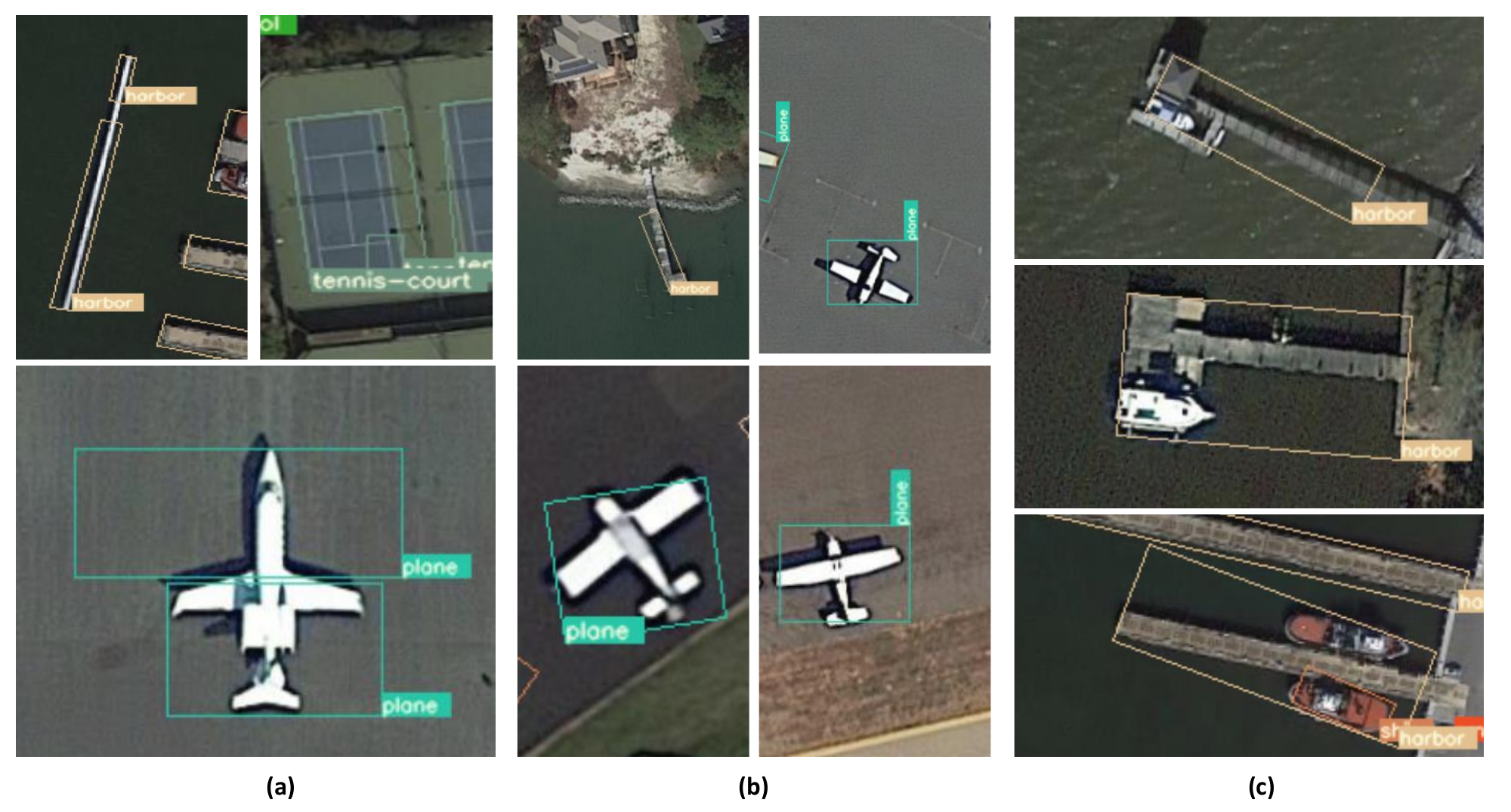}
\end{center}
   \caption{Sample failed detection results of our proposed method on the DOTA dataset. (a) Noisy or incorrectly detected bounding boxes. (b) Bounding boxes not aligned with the major axes of the objects. (c) A few rare failed cases where smaller objects inside larger objects have not been detected.}
\label{fig:failure}
\end{figure*}

\subsection{Limitations}
Fig.~\ref{fig:failure} represents some failure cases on OBB task of DOTA dataset. There are three main types of mistakes in the detection results, although some are negligible. The first type occurs in  categories with long object instances,
as illustrated in Fig.~\ref{fig:failure}(a) as some failures in the \emph{harbor}, \emph{tennis-court}, and \emph{plane} classes. Although the class labels are correctly recognized, the bounding boxes may split into two parts or include smaller artifact boxes. Some of these failures,
such as
incorrectly detecting the bounding box of the \emph{plane}, however, are related to the random image splitting whereby some objects are divided into two separate images. The second type of mistake, which is also negligible, happens when detected boxes are not tightly aligned with the major axes of the objects. As shown in Fig.~\ref{fig:failure}(b), however, the bounding boxes still surround the objects. 
Another rare case occurs when our method fails to detect smaller objects which occur inside other larger objects. An example of this is shown in Fig.~\ref{fig:failure}(c) where \emph{ships}  are not recognized in  \emph{harbors}, and the detections favor the larger objects (i.e. \emph{harbor}).

\section{Conclusion}
This paper presents a novel object detection method to deal with the main challenges encountered in remotely sensed imagery. Localizing tiny objects 
over large areas requires extremely dense grids with efficient bounding box proposal networks, which existing methods lack. The proposed method achieves this using an effective approach that 
allows the object pixels to participate
directly in the bounding box prediction. It thereby leverages the advantages of high resolution remote sensing images by 
utilizing more local details of the objects on the ground, while high-level features and object-level information are also used within a larger receptive field
in
subsequent processing steps. 
Specifically, 
the method casts the regression task for predicting bounding boxes into a pure classification problem, exploiting
the benefits of
discrete domain classification,
which is known to be less error prone than
the alternative continuous domain regression. This also enables detecting oriented bounding boxes without any extra computation. One important advantage of the proposed method is the lack of dependency on IoU (intersection over union) and anchor box selection, since they are already selected based on the object properties in the image. In addition, rotation-invariant feature learning is used to enforce multiple rotated samples that share similar features via a regularization constraint. 

While motivated by the specific  challenge of identifying small objects in remote sensing images,  the proposed method does not require any specific knowledge about the dataset or pre-defined anchor boxes. It can therefore potentially be generalized to other applications as well, where detecting small objects is of interest. Moreover, the object orientation obtained from the oriented bounding boxes can be used in other vision problems such as object tracking or action recognition.  In the xView dataset, tiny objects are fine grained and belong to different parent classes. For instance, 8 distinct truck classes can be seen as a single class. A hierarchical feature extractor therefore might be beneficial to further enhance the results.


%





\ifCLASSOPTIONcaptionsoff
  \newpage
\fi



\bibliographystyle{IEEEtran}
\bibliography{IEEEabrv,refs}

\begin{thebibliography}{10}
\providecommand{\url}[1]{#1}
\csname url@samestyle\endcsname
\providecommand{\newblock}{\relax}
\providecommand{\bibinfo}[2]{#2}
\providecommand{\BIBentrySTDinterwordspacing}{\spaceskip=0pt\relax}
\providecommand{\BIBentryALTinterwordstretchfactor}{4}
\providecommand{\BIBentryALTinterwordspacing}{\spaceskip=\fontdimen2\font plus
\BIBentryALTinterwordstretchfactor\fontdimen3\font minus
  \fontdimen4\font\relax}
\providecommand{\BIBforeignlanguage}[2]{{%
\expandafter\ifx\csname l@#1\endcsname\relax
\typeout{** WARNING: IEEEtran.bst: No hyphenation pattern has been}%
\typeout{** loaded for the language `#1'. Using the pattern for}%
\typeout{** the default language instead.}%
\else
\language=\csname l@#1\endcsname
\fi
#2}}
\providecommand{\BIBdecl}{\relax}
\BIBdecl

\bibitem{liu2016ship}
Z.~Liu, H.~Wang, L.~Weng, and Y.~Yang, ``Ship rotated bounding box space for
  ship extraction from high-resolution optical satellite images with complex
  backgrounds,'' \emph{IEEE Geoscience and Remote Sensing Letters}, vol.~13,
  no.~8, pp. 1074--1078, 2016.

\bibitem{kalantar2017multiple}
B.~Kalantar, S.~B. Mansor, A.~A. Halin, H.~Z.~M. Shafri, and M.~Zand,
  ``Multiple moving object detection from uav videos using trajectories of
  matched regional adjacency graphs,'' \emph{IEEE Transactions on Geoscience
  and Remote Sensing}, vol.~55, no.~9, pp. 5198--5213, 2017.

\bibitem{li2020object}
K.~Li, G.~Wan, G.~Cheng, L.~Meng, and J.~Han, ``Object detection in optical
  remote sensing images: A survey and a new benchmark,'' \emph{ISPRS Journal of
  Photogrammetry and Remote Sensing}, vol. 159, pp. 296--307, 2020.

\bibitem{usman2019weakly}
M.~Usman~Rafique, H.~Blanton, and N.~Jacobs, ``Weakly supervised fusion of
  multiple overhead images,'' in \emph{Proceedings of the IEEE Conference on
  Computer Vision and Pattern Recognition Workshops}, 2019, pp. 0--0.

\bibitem{khan2017forest}
S.~H. Khan, X.~He, F.~Porikli, and M.~Bennamoun, ``Forest change detection in
  incomplete satellite images with deep neural networks,'' \emph{IEEE
  Transactions on Geoscience and Remote Sensing}, vol.~55, no.~9, pp.
  5407--5423, 2017.

\bibitem{ren2015faster}
S.~Ren, K.~He, R.~Girshick, and J.~Sun, ``Faster r-cnn: Towards real-time
  object detection with region proposal networks,'' in \emph{Advances in neural
  information processing systems}, 2015, pp. 91--99.

\bibitem{redmon2018yolov3}
J.~Redmon and A.~Farhadi, ``Yolov3: An incremental improvement,'' \emph{arXiv
  preprint arXiv:1804.02767}, 2018.

\bibitem{girshick2014rich}
R.~Girshick, J.~Donahue, T.~Darrell, and J.~Malik, ``Rich feature hierarchies
  for accurate object detection and semantic segmentation,'' in
  \emph{Proceedings of the IEEE conference on computer vision and pattern
  recognition}, 2014, pp. 580--587.

\bibitem{kemker2018algorithms}
R.~Kemker, C.~Salvaggio, and C.~Kanan, ``Algorithms for semantic segmentation
  of multispectral remote sensing imagery using deep learning,'' \emph{ISPRS
  journal of photogrammetry and remote sensing}, vol. 145, pp. 60--77, 2018.

\bibitem{van2018you}
A.~Van~Etten, ``You only look twice: Rapid multi-scale object detection in
  satellite imagery,'' \emph{arXiv preprint arXiv:1805.09512}, 2018.

\bibitem{xia2018dota}
G.-S. Xia, X.~Bai, J.~Ding, Z.~Zhu, S.~Belongie, J.~Luo, M.~Datcu, M.~Pelillo,
  and L.~Zhang, ``Dota: A large-scale dataset for object detection in aerial
  images,'' in \emph{Proceedings of the IEEE Conference on Computer Vision and
  Pattern Recognition}, 2018, pp. 3974--3983.

\bibitem{azimi2018towards}
S.~M. Azimi, E.~Vig, R.~Bahmanyar, M.~K{\"o}rner, and P.~Reinartz, ``Towards
  multi-class object detection in unconstrained remote sensing imagery,'' in
  \emph{Asian Conference on Computer Vision}.\hskip 1em plus 0.5em minus
  0.4em\relax Springer, 2018, pp. 150--165.

\bibitem{yang2019scrdet}
X.~Yang, J.~Yang, J.~Yan, Y.~Zhang, T.~Zhang, Z.~Guo, X.~Sun, and K.~Fu,
  ``Scrdet: Towards more robust detection for small, cluttered and rotated
  objects,'' in \emph{Proceedings of the IEEE International Conference on
  Computer Vision}, 2019, pp. 8232--8241.

\bibitem{neubeck2006efficient}
A.~Neubeck and L.~Van~Gool, ``Efficient non-maximum suppression,'' in
  \emph{18th International Conference on Pattern Recognition (ICPR'06)},
  vol.~3.\hskip 1em plus 0.5em minus 0.4em\relax IEEE, 2006, pp. 850--855.

\bibitem{zhang2019scale}
S.~Zhang, G.~He, H.-B. Chen, N.~Jing, and Q.~Wang, ``Scale adaptive proposal
  network for object detection in remote sensing images,'' \emph{IEEE
  Geoscience and Remote Sensing Letters}, vol.~16, no.~6, pp. 864--868, 2019.

\bibitem{zhuang2019single}
S.~Zhuang, P.~Wang, B.~Jiang, G.~Wang, and C.~Wang, ``A single shot framework
  with multi-scale feature fusion for geospatial object detection,''
  \emph{Remote Sensing}, vol.~11, no.~5, p. 594, 2019.

\bibitem{qiu2019a2rmnet}
H.~Qiu, H.~Li, Q.~Wu, F.~Meng, K.~N. Ngan, and H.~Shi, ``A2rmnet: Adaptively
  aspect ratio multi-scale network for object detection in remote sensing
  images,'' \emph{Remote Sensing}, vol.~11, no.~13, p. 1594, 2019.

\bibitem{li2018multiscale}
S.~Li, Z.~Zhang, B.~Li, and C.~Li, ``Multiscale rotated bounding box-based deep
  learning method for detecting ship targets in remote sensing images,''
  \emph{Sensors}, vol.~18, no.~8, p. 2702, 2018.

\bibitem{liu2016ssd}
W.~Liu, D.~Anguelov, D.~Erhan, C.~Szegedy, S.~Reed, C.-Y. Fu, and A.~C. Berg,
  ``Ssd: Single shot multibox detector,'' in \emph{European conference on
  computer vision}.\hskip 1em plus 0.5em minus 0.4em\relax Springer, 2016, pp.
  21--37.

\bibitem{sergievskiy2019reduced}
N.~Sergievskiy and A.~Ponamarev, ``Reduced focal loss: 1st place solution to
  xview object detection in satellite imagery,'' \emph{arXiv preprint
  arXiv:1903.01347}, 2019.

\bibitem{yang2018automatic}
X.~Yang, H.~Sun, K.~Fu, J.~Yang, X.~Sun, M.~Yan, and Z.~Guo, ``Automatic ship
  detection in remote sensing images from google earth of complex scenes based
  on multiscale rotation dense feature pyramid networks,'' \emph{Remote
  Sensing}, vol.~10, no.~1, p. 132, 2018.

\bibitem{ma2018arbitrary}
J.~Ma, W.~Shao, H.~Ye, L.~Wang, H.~Wang, Y.~Zheng, and X.~Xue,
  ``Arbitrary-oriented scene text detection via rotation proposals,''
  \emph{IEEE Transactions on Multimedia}, vol.~20, no.~11, pp. 3111--3122,
  2018.

\bibitem{ding2019learning}
J.~Ding, N.~Xue, Y.~Long, G.-S. Xia, and Q.~Lu, ``Learning roi transformer for
  oriented object detection in aerial images,'' in \emph{Proceedings of the
  IEEE Conference on Computer Vision and Pattern Recognition}, 2019, pp.
  2849--2858.

\bibitem{liu2017high}
Z.~Liu, L.~Yuan, L.~Weng, and Y.~Yang, ``A high resolution optical satellite
  image dataset for ship recognition and some new baselines,'' in
  \emph{International Conference on Pattern Recognition Applications and
  Methods}, vol.~2.\hskip 1em plus 0.5em minus 0.4em\relax SCITEPRESS, 2017,
  pp. 324--331.

\bibitem{li2017rotation}
K.~Li, G.~Cheng, S.~Bu, and X.~You, ``Rotation-insensitive and
  context-augmented object detection in remote sensing images,'' \emph{IEEE
  Transactions on Geoscience and Remote Sensing}, vol.~56, no.~4, pp.
  2337--2348, 2017.

\bibitem{cheng2016learning}
G.~Cheng, P.~Zhou, and J.~Han, ``Learning rotation-invariant convolutional
  neural networks for object detection in vhr optical remote sensing images,''
  \emph{IEEE Transactions on Geoscience and Remote Sensing}, vol.~54, no.~12,
  pp. 7405--7415, 2016.

\bibitem{guo2018geospatial}
W.~Guo, W.~Yang, H.~Zhang, and G.~Hua, ``Geospatial object detection in high
  resolution satellite images based on multi-scale convolutional neural
  network,'' \emph{Remote Sensing}, vol.~10, no.~1, p. 131, 2018.

\bibitem{he2016deep}
K.~He, X.~Zhang, S.~Ren, and J.~Sun, ``Deep residual learning for image
  recognition,'' in \emph{Proceedings of the IEEE conference on computer vision
  and pattern recognition}, 2016, pp. 770--778.

\bibitem{zhu2015orientation}
H.~Zhu, X.~Chen, W.~Dai, K.~Fu, Q.~Ye, and J.~Jiao, ``Orientation robust object
  detection in aerial images using deep convolutional neural network,'' in
  \emph{2015 IEEE International Conference on Image Processing (ICIP)}.\hskip
  1em plus 0.5em minus 0.4em\relax IEEE, 2015, pp. 3735--3739.

\bibitem{mundhenk2016large}
T.~N. Mundhenk, G.~Konjevod, W.~A. Sakla, and K.~Boakye, ``A large contextual
  dataset for classification, detection and counting of cars with deep
  learning,'' in \emph{European Conference on Computer Vision}.\hskip 1em plus
  0.5em minus 0.4em\relax Springer, 2016, pp. 785--800.

\bibitem{zhang2019cad}
G.~Zhang, S.~Lu, and W.~Zhang, ``Cad-net: A context-aware detection network for
  objects in remote sensing imagery,'' \emph{IEEE Transactions on Geoscience
  and Remote Sensing}, vol.~57, no.~12, pp. 10\,015--10\,024, 2019.

\bibitem{wang2019sard}
Y.~Wang, Y.~Zhang, Y.~Zhang, L.~Zhao, X.~Sun, and Z.~Guo, ``Sard: Towards
  scale-aware rotated object detection in aerial imagery,'' \emph{IEEE Access},
  vol.~7, pp. 173\,855--173\,865, 2019.

\bibitem{lam2018xview}
D.~Lam, R.~Kuzma, K.~McGee, S.~Dooley, M.~Laielli, M.~Klaric, Y.~Bulatov, and
  B.~McCord, ``xview: Objects in context in overhead imagery,'' \emph{arXiv
  preprint arXiv:1802.07856}, 2018.

\bibitem{wei2019oriented}
H.~Wei, L.~Zhou, Y.~Zhang, H.~Li, R.~Guo, and H.~Wang, ``Oriented objects as
  pairs of middle lines,'' \emph{arXiv preprint arXiv:1912.10694}, 2019.

\bibitem{law2018cornernet}
H.~Law and J.~Deng, ``Cornernet: Detecting objects as paired keypoints,'' in
  \emph{Proceedings of the European Conference on Computer Vision (ECCV)},
  2018, pp. 734--750.

\bibitem{long2017accurate}
Y.~Long, Y.~Gong, Z.~Xiao, and Q.~Liu, ``Accurate object localization in remote
  sensing images based on convolutional neural networks,'' \emph{IEEE
  Transactions on Geoscience and Remote Sensing}, vol.~55, no.~5, pp.
  2486--2498, 2017.

\bibitem{zhou2020objects}
L.~Zhou, H.~Wei, H.~Li, Y.~Zhang, X.~Sun, and W.~Zhao, ``Objects detection for
  remote sensing images based on polar coordinates,'' \emph{arXiv preprint
  arXiv:2001.02988}, 2020.

\bibitem{girshick2015fast}
R.~Girshick, ``Fast r-cnn,'' in \emph{Proceedings of the IEEE international
  conference on computer vision}, 2015, pp. 1440--1448.

\bibitem{redmon2016you}
J.~Redmon, S.~Divvala, R.~Girshick, and A.~Farhadi, ``You only look once:
  Unified, real-time object detection,'' in \emph{Proceedings of the IEEE
  conference on computer vision and pattern recognition}, 2016, pp. 779--788.

\bibitem{redmon2017yolo9000}
J.~Redmon and A.~Farhadi, ``Yolo9000: better, faster, stronger,'' in
  \emph{Proceedings of the IEEE conference on computer vision and pattern
  recognition}, 2017, pp. 7263--7271.

\bibitem{cheng2016survey}
G.~Cheng and J.~Han, ``A survey on object detection in optical remote sensing
  images,'' \emph{ISPRS Journal of Photogrammetry and Remote Sensing}, vol.
  117, pp. 11--28, 2016.

\bibitem{cheng2020cross}
G.~Cheng, Y.~Si, H.~Hong, X.~Yao, and L.~Guo, ``Cross-scale feature fusion for
  object detection in optical remote sensing images,'' \emph{IEEE Geoscience
  and Remote Sensing Letters}, 2020.

\bibitem{lin2017feature}
T.-Y. Lin, P.~Doll{\'a}r, R.~Girshick, K.~He, B.~Hariharan, and S.~Belongie,
  ``Feature pyramid networks for object detection,'' in \emph{Proceedings of
  the IEEE conference on computer vision and pattern recognition}, 2017, pp.
  2117--2125.

\bibitem{tian2019fcos}
Z.~Tian, C.~Shen, H.~Chen, and T.~He, ``Fcos: Fully convolutional one-stage
  object detection,'' in \emph{Proceedings of the IEEE international conference
  on computer vision}, 2019, pp. 9627--9636.

\bibitem{yang2018multi}
T.~Yang, S.~Qin, J.~Yan, and W.~Zhang, ``Multi-label dilated recurrent network
  for sequential face alignment,'' in \emph{2018 IEEE International Conference
  on Multimedia and Expo (ICME)}.\hskip 1em plus 0.5em minus 0.4em\relax IEEE,
  2018, pp. 1--6.

\bibitem{liu2020rgb}
Y.~Liu, H.~Xiao, H.~Tan, and P.~Li, ``Are rgb-based salient object detection
  methods unsuitable for light field data?'' \emph{EURASIP Journal on Image and
  Video Processing}, vol. 2020, no.~1, pp. 1--17, 2020.

\bibitem{szegedy2015going}
C.~Szegedy, W.~Liu, Y.~Jia, P.~Sermanet, S.~Reed, D.~Anguelov, D.~Erhan,
  V.~Vanhoucke, and A.~Rabinovich, ``Going deeper with convolutions,'' in
  \emph{Proceedings of the IEEE conference on computer vision and pattern
  recognition}, 2015, pp. 1--9.

\bibitem{deng2018multi}
Z.~Deng, H.~Sun, S.~Zhou, J.~Zhao, L.~Lei, and H.~Zou, ``Multi-scale object
  detection in remote sensing imagery with convolutional neural networks,''
  \emph{ISPRS journal of photogrammetry and remote sensing}, vol. 145, pp.
  3--22, 2018.

\bibitem{tayara2018object}
H.~Tayara and K.~T. Chong, ``Object detection in very high-resolution aerial
  images using one-stage densely connected feature pyramid network,''
  \emph{Sensors}, vol.~18, no.~10, p. 3341, 2018.

\bibitem{cheng2016rifd}
G.~Cheng, P.~Zhou, and J.~Han, ``Rifd-cnn: Rotation-invariant and fisher
  discriminative convolutional neural networks for object detection,'' in
  \emph{Proceedings of the IEEE conference on computer vision and pattern
  recognition}, 2016, pp. 2884--2893.

\bibitem{cheng2018learning}
G.~Cheng, J.~Han, P.~Zhou, and D.~Xu, ``Learning rotation-invariant and fisher
  discriminative convolutional neural networks for object detection,''
  \emph{IEEE Transactions on Image Processing}, vol.~28, no.~1, pp. 265--278,
  2018.

\bibitem{duan2019large}
L.~Duan, M.~Desbrun, A.~Giraud, F.~Trastour, and L.~Laurore, ``Large-scale dtm
  generation from satellite data,'' in \emph{Proceedings of the IEEE Conference
  on Computer Vision and Pattern Recognition Workshops}, 2019, pp. 0--0.

\bibitem{suzuki1985topological}
S.~Suzuki \emph{et~al.}, ``Topological structural analysis of digitized binary
  images by border following,'' \emph{Computer vision, graphics, and image
  processing}, vol.~30, no.~1, pp. 32--46, 1985.

\bibitem{sklansky1982finding}
J.~Sklansky, ``Finding the convex hull of a simple polygon,'' \emph{Pattern
  Recognition Letters}, vol.~1, no.~2, pp. 79--83, 1982.

\bibitem{toussaint1983solving}
G.~T. Toussaint, ``Solving geometric problems with the rotating calipers,'' in
  \emph{Proc. IEEE Melecon}, vol.~83, 1983, p. A10.

\bibitem{bradski2008learning}
G.~Bradski and A.~Kaehler, \emph{Learning OpenCV: Computer vision with the
  OpenCV library}.\hskip 1em plus 0.5em minus 0.4em\relax " O'Reilly Media,
  Inc.", 2008.

\bibitem{simonyan2014very}
K.~Simonyan and A.~Zisserman, ``Very deep convolutional networks for
  large-scale image recognition,'' \emph{arXiv preprint arXiv:1409.1556}, 2014.

\bibitem{ronneberger2015u}
O.~Ronneberger, P.~Fischer, and T.~Brox, ``U-net: Convolutional networks for
  biomedical image segmentation,'' in \emph{International Conference on Medical
  image computing and computer-assisted intervention}.\hskip 1em plus 0.5em
  minus 0.4em\relax Springer, 2015, pp. 234--241.

\bibitem{badrinarayanan2017segnet}
V.~Badrinarayanan, A.~Kendall, and R.~Cipolla, ``Segnet: A deep convolutional
  encoder-decoder architecture for image segmentation,'' \emph{IEEE
  transactions on pattern analysis and machine intelligence}, vol.~39, no.~12,
  pp. 2481--2495, 2017.

\bibitem{li2020scattnet}
H.~Li, K.~Qiu, L.~Chen, X.~Mei, L.~Hong, and C.~Tao, ``Scattnet: Semantic
  segmentation network with spatial and channel attention mechanism for
  high-resolution remote sensing images,'' \emph{IEEE Geoscience and Remote
  Sensing Letters}, 2020.

\end{thebibliography}
\end{document}